\documentclass[twocolumn]{IEEEtran}

 \usepackage{epsfig,graphicx}
 \usepackage{supertabular}
 \usepackage{amsmath}
 \usepackage{amsxtra}
 \usepackage{dblfloatfix}
 \usepackage{amssymb}
 \usepackage{graphics}
 
 \providecommand{\norm}[1]{{\lVert#1\rVert}}

\begin{document}

\title{Distributed Representation of Geometrically Correlated Images with Compressed Linear Measurements
\thanks{This work
has been partly supported by the Swiss National Science Foundation,
under grant 200021-118230. This work was presented  (in part) at the IEEE International Conference on Acoustics, Speech and Signal Processing (ICASSP), Dallas, March 2010 \cite{Vijay_ICASSP2010}, and at the European Signal Processing Conference (EUSIPCO), Aalborg, Denmark, Aug. 2010  \cite{Vijay_EUSIPCO2010}. }}

\author{Vijayaraghavan~Thirumalai and Pascal~Frossard \\
        Ecole Polytechnique F\'ed\'erale de Lausanne (EPFL) \\ Signal Processing Laboratory (LTS4) , Lausanne, 1015 - Switzerland.
        \\ Email:\{vijayaraghavan.thirumalai, pascal.frossard\}@epfl.ch \\
        Fax: +41 21 693 7600, Phone: +41 21 693 2708}

\maketitle 

\begin{abstract}

This paper addresses the problem of distributed coding of images whose correlation is driven by the motion of objects or positioning of the vision sensors. It concentrates on the problem where images are encoded with compressed linear measurements. We propose a geometry-based correlation model in order to describe the common information in pairs of images. We assume that the constitutive components of natural images can be captured by visual features that undergo local transformations (e.g., translation) in different images. We first identify prominent visual features by computing a sparse approximation of a reference image with a dictionary of geometric basis functions. We then pose a regularized optimization problem to estimate the corresponding features in correlated images given by quantized linear measurements. The estimated features have to comply with the compressed information and to represent consistent transformation between images. The correlation model is given by the relative geometric transformations between corresponding features. We then propose an efficient joint decoding algorithm that estimates the compressed images such that they stay consistent with both the quantized measurements and the correlation model. Experimental results show that the proposed algorithm effectively estimates the correlation between images in multi-view datasets. In addition, the proposed algorithm provides effective decoding performance that compares advantageously to independent coding solutions as well as state-of-the-art distributed coding schemes based on disparity learning.
\end{abstract}


\section{Introduction}

\IEEEPARstart{I}n recent years, vision sensor networks have been gaining an ever increasing popularity enforced by the availability of cheap semiconductor components. These networks typically produce  highly redundant information so that an efficient estimation of the correlation between images becomes primordial for effective coding, transmission and storage applications. The distributed coding paradigm becomes particularly attractive in such settings; it permits to efficiently exploit the correlation between images with low encoding complexity and minimal inter-sensor communication, which translates into power savings in sensor networks. One of the most important challenging tasks however resides in the proper modeling and estimation of the correlation between images. 

In this paper, we consider the problem of finding an efficient distributed representation for correlated images, where the common objects are displaced due to the viewpoint changes or motion in dynamic scenes. In particular, we are interested in a scenario where the images are given under the form of few quantized linear measurements computed by very simple sensors. Even with such a simple acquisition stage, the images can be reconstructed under the condition that they have a sparse representation in particular basis (e.g., DCT, wavelet) that is sufficiently different from the sensing matrices \cite{Donoho,Candes}. Rather than independent image reconstruction, we are however interested in the joint reconstruction of the images and in particular the estimation of their correlation from the compressed measurements. In contrary to most distributed compressive schemes in the literature, we want to estimate the correlation prior to image reconstruction for improved robustness at low coding rates. 

We propose to model the correlation between images as geometric transformations of visual features, which provides a more efficient representation than block-based translational models that are commonly used in state-of-the-art coding solutions. We first compute the most prominent visual features in a reference image through a sparse approximation with geometric functions drawn from a parametric dictionary. Then, we formulate a regularized optimization problem whose objective is to identify in the compressed images the features that correspond to the prominent components in the reference images. Correspondences then define relative transformations between images that form the geometric correlation model. A regularization constraint ensures that the estimated correlation is consistent and corresponds to the actual motion of visual objects. We then use the estimated correlation in a new joint decoding algorithm that approximates the multiple images. The joint decoding is cast as an optimization problem that warps the reference image according to the transformation described in the correlation information, while enforcing the decoded images to be consistent with the quantized measurements. We finally propose an extension of our algorithm to the joint decoding of multi-view images.

While our novel framework could find applications in several problems such as distributed video coding or multi-view imaging, we focus on the latter for illustrating the joint decoding performance. We show by experiments that the proposed algorithm computes a good estimation of the correlation between multi-view images. In particular, the results confirm that dictionaries based on geometric basis functions permit to capture the correlation more efficiently than a dictionary built on patches or blocks from the reference image \cite{YiMa_PCS}. In addition, we show that the estimated correlation model can be used to decode the compressed image by disparity compensation. Such a decoding strategy permits to outperform independent coding solutions based on JPEG~2000 and state-of-the-art distributed coding schemes based on disparity learning \cite{David,David1} in terms of rate-distortion (RD) performance due to accurate correlation estimation. Finally, the experiments outline that enforcing consistency in image prediction is very effective in increasing the decoding quality when the images are given by the quantized linear measurements.

The rest of this paper is organized as follows. Section \ref{sec:rel_work} briefly overviews the related work with a emphasis on reconstruction from random projections and distributed coding algorithms. The geometric correlation model used in our framework is presented in Section \ref{sec:framework}. Section \ref{sec:estimating_motion} describes the proposed regularized energy model for an image pair and the optimization algorithm. The consistent image prediction algorithm is described in Section \ref{sec:cons_pred}. Section \ref{sec:ext_mv} describes the extension of our scheme to multi-view images. Finally, experimental results are presented in Section \ref{sec:Results} and Section \ref{sec:conc} concludes this paper.

\section{Related work} \label{sec:rel_work}

We present in this section a brief overview of the related works in distributed image coding, where we mostly focus on simple sensing solutions based on linear measurements. In recent years, signal acquisition based on random projections has actually received a significant attention in many applications like medical imaging, compressive imaging or sensor networks. Donoho \cite{Donoho} and Candes \emph{et al.} \cite{Candes} have shown that a small number of linear measurements contain enough information to reconstruct a signal, as long as it has sparse representation in a basis that is incoherent with the sensing matrix \cite{Candes_imagecoding}. Rauhut \emph{et al.} \cite{Rauhut} extend the concept of signal reconstruction from linear measurements using redundant dictionaries. Signal reconstruction from linear measurements has been applied to different applications such as image acquisition \cite{singlepixel_CS, Mun_blockCS, Gan_Eusipco} and video representation \cite{Stankovic,Park,Vaswani}.

At the same time, the key in effective distributed representation certainly lies in the definition of good correlation models. Duarte \emph{et al.} \cite{Duarte_DCS,Duarte_DCS1} have proposed different correlation models for the distributed compression of correlated signals from linear measurements. In particular, they introduce three joint sparsity models (JSM) in order to exploit the inter-signal correlation in the joint reconstruction. These three sparse models are respectively described by (i) JSM-1, where the signals share a common sparse support plus a sparse innovation part specific to each signal; (ii) JSM-2, where the signals share a common sparse support with different coefficients, and (iii) JSM-3 with a non-sparse common signal with individual sparse innovation in each signal. These correlation models permit a joint reconstruction with a reduced sampling rate or equivalently a smaller number of measurements compared to independent reconstruction for the same decoding quality. The sparsity models developed in \cite{Duarte_DCS} have then been applied to distributed video coding \cite{Kang_ICASSP,Do_ICIP} with random projections. The scheme in \cite{Kang_ICASSP} used a modified gradient projection sparse algorithm \cite{GPSR} for the joint signal reconstruction. The authors in \cite{Do_ICIP} have proposed a distributed compressive video coding scheme based on the sparse recovery with decoder side information. In particular, the prediction error between the original and side information frames is assumed to be sparse in a particular orthonormal basis (e.g., wavelet basis). Another distributed video coding scheme has been proposed in \cite{YiMa_PCS}, which relies on an inter-frame sparsity model. A block of pixels in a frame is assumed to be sparsely represented by linear combination of the neighboring blocks from the decoded key frames. In particular, an adaptive block-based dictionary is constructed from the previously decoded key frames and eventually used for signal reconstruction. Finally, iterative projection methods are used in \cite{Trocan_ICME,Trocan_MMSP} in order to ensure a joint reconstruction of correlated images that are sparse in a dual tree wavelet transform basis and at the same time consistent with the linear measurements in multi-view settings.
In general, state-of-the-art distributed compressive schemes \cite{Kang_ICASSP,Do_ICIP,Trocan_ICME,Trocan_MMSP} estimates the correlation model from two reconstructed reference images, where the reference frames are reconstructed from the respective linear measurements by solving an $l_2$-TV or $l_2$-$l_1$ optimization problem. Unfortunately, reconstructing the reference images based on solving an $l_2$-$l_1$ or $l_2$-TV optimization problem is computationally expensive \cite{Donoho,Candes}. Also, the correlation model estimated from highly compressed reference images usually fails to capture the actual geometrical relationship between images. Motivated by these issues, we  estimate in this paper a robust correlation model directly from the highly compressed linear measurements using a reference image, without explicitly reconstructing the compressed images.

\begin{figure*}
\begin{minipage}[b]{1.0\linewidth}
 \centering
 \centerline{\epsfig{figure=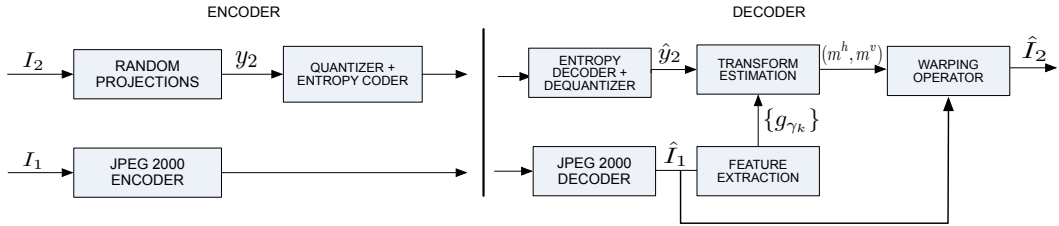,width=14.5cm}}
\end{minipage}
\caption{Schematic representation of the proposed scheme. The images $I_1$ and $I_2$ are correlated through displacement of scene objects due to viewpoint change.}
\label{Fig:block_scheme}
\end{figure*}

In multi-view imaging or distributed video coding, the correlation is explained by the motion of objects or the change of viewpoint. Block-based translation models that are commonly used for correlation estimation fail to efficiently capture the geometry of objects. This results in poor correlation model especially with low resolution images. Furthermore, most of the above mentioned schemes (except \cite{YiMa_PCS}) assume that the signal is sparse in a particular orthonormal basis (e.g., DCT or Wavelet). This is also the case of the JSM models described above which cannot be used to relate the scene objects by means of a local transform, and unfortunately fail to provide an efficient joint representation of correlated images at low bit rates. It is more generic to assume the signals to be sparse in a redundant dictionary which allows greater flexibility in the design of the representation vectors. The most prominent geometric components in the images can be captured efficiently by dictionary functions. The correlation can be then estimated by comparing the most prominent features in different images. Few works have been reported in the literature for the estimation of a correlation model using redundant structured dictionaries in multi-view \cite{Ivana_TIP} or video applications \cite{Oscar_TIP}.  However, these frameworks do not construct the correlation model from the linear measurements. In general, most of the schemes in classical disparity and motion estimation focus on estimating correlation from original images \cite{Scharstein,Baker_flow}, and not from compressed images. We rather focus here on estimating the correlation from compressed images where the image is given with random linear measurements. The correlation model is built using the geometric transformations captured by a structured dictionary which leads to an effective estimation of the geometric correlation between images.

Finally, the distributed schemes in the literature that are based on compressed measurements usually fail to estimate the actual number of bits for the image sequence representation (except \cite{YiMa_PCS}), and hence cannot be applied directly in practical coding applications. Quantization and entropy coding of the measurements is actually an open research problem due to the two following reasons: (i) the reconstructed signal from quantized measurements does not necessarily satisfy the consistent reconstruction property \cite{one_bit_CS}; (ii) the entropy of the measurements is usually large which leads to unsatisfactory coding performance in imaging applications \cite{Schulz}. Hence, it is essential to adapt the quantization techniques and reconstruction algorithms in order to reduce the distortion in the reconstructed signal such as \cite{Laurent_BPDNp,Zymnis}. The authors in \cite{Fletcher,Dai} have also studied the asymptotic reconstruction performance of the signal under uniform and non-uniform quantization schemes. They have shown that a non-uniform quantization scheme usually gives smaller distortion in the reconstruction signal comparing to a uniform quantization scheme. Recently, optimal quantization strategy for the random measurements has been designed based on distributed functional scalar quantizers \cite{sun_ISIT}. In this paper, we use a simple quantization strategy for realistic compression along with consistent prediction constraints in the joint decoding of correlated images in order to illustrate the potential of low complexity sensing solutions in practical multi-view imaging applications. 

%

\section{Framework} \label{sec:framework}

 We consider a pair of images $I_1$ and $I_2$ (with resolution $N = N_1 \times N_2$) that represent a scene taken from different viewpoints; these images are correlated through motion of visual objects. The captured images are encoded independently and are transmitted to a joint decoder. The joint decoder estimates the relative transformations between the received signals and jointly decodes the images. While the description is given here for pairs of images, we later extend the framework to multiple images in Section \ref{sec:ext_mv}. 

We focus on the particular problem where one of the images serves as a reference for the correlation estimation and the decoding of second image as illustrated in Fig.~\ref{Fig:block_scheme}. While the reference image $I_1$ could be encoded with any compression algorithm (e.g., JPEG, compressed sensing framework \cite{Gan_Eusipco}), we choose here to encode the reference image ${I}_1$ with JPEG~2000 coding solutions. Next, we concentrate on the independent coding and joint decoding of the second image, where the first image $\hat{I}_1$ serves as side information. The second image $I_2$ is projected on a random matrix $\Phi$ to generate the measurements $y_2 = \Phi I_2$. The measurements $y_2$ are quantized with a uniform quantization algorithm and the quantized linear measurements are finally compressed with an entropy coder.

At the decoder, we first estimate the prominent visual features that carry the geometry information of the objects in the scene. In particular, the decoder computes a sparse approximation of the image $\hat{I}_1$ using a parametric dictionary of geometric functions. Such an approximation captures the most prominent geometrical features in the image $\hat{I}_1$. We then estimate the corresponding features in the second image $I_2$ directly from the quantized linear measurements $\hat{y}_2$ without explicit image reconstruction. In particular, the corresponding features between images are related using a geometry-based correlation model, where the correspondences describe local geometric transformations between images. The correlation information is further used to decode the compressed image $\hat{I}_2$ from the reference image $\hat{I}_1$. We finally ensure a consistent prediction of $\hat{I}_2$ by explicitly considering the quantized measurements $\hat{y}_2$ during the warping process. Before getting into the details of the correlation estimation algorithm, we describe the sparse  approximation algorithm and the geometry-based correlation model built on a parametric dictionary.

\begin{figure}
 \centering
$\begin{array}{@{\hspace{0.1in}} c@{\hspace{0.3in}}c @{\hspace{0.3in}} c} \multicolumn{1}{l}{\mbox{}} &  \multicolumn{1}{l}{\mbox{}} &\multicolumn{1}{l}{\mbox{}} \\
   \epsfxsize=0.7in \epsffile{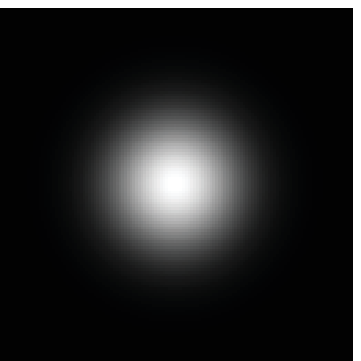} & \epsfxsize=0.7in \epsffile{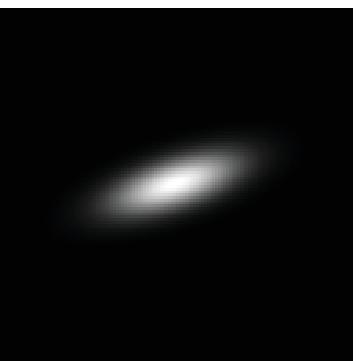} & \epsfxsize=0.7in \epsffile{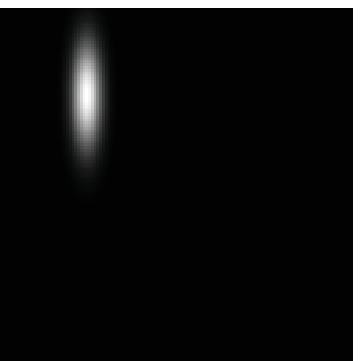} \\ 
     \mbox{(a) } & \mbox{(b) } & \mbox{(c) }   
   \end{array}$
\caption{Sample Gaussian atoms with mother function ${g(x,y) = \frac{1}{\sqrt{\pi}}\mbox{exp}(-(x^2+y^2))}$ that undergo different set of transformations.}
  \label{Fig:gaussian_atoms}
  \end{figure}

We describe now the geometric correlation model that is based on matching the sparse geometric features in different images. We first compute a sparse approximation of the reference image $\hat{I}_1$ using geometric basis functions in a structured dictionary $\mathcal{D} = \{ g_{\gamma} \}$ where $g_\gamma$ is called an \emph{atom}. The dictionary $\mathcal{D}$ is typically constructed by applying geometric transformations (given by the unitary operator $U(\gamma)$) to a generating function $g$ to form the atom $g_\gamma$. A geometric transformation indexed by $\gamma$ consists of a combination of operators for anisotropic scale by $s_x,s_y$, rotation by $\theta$, and translation by $t_x,t_y$. For example, when $g$ is a Gaussian function $g(x,y) = \frac{1}{\sqrt{\pi}}\mbox{exp}(-(x^2+y^2))$, the transformation $g_{\gamma}$ is expressed as  
\begin{eqnarray}
g_{\gamma}(x,y) = \frac{1}{\sqrt{\pi}}\mbox{exp}(-(g_1^2+g_2^2)) \\ \nonumber
\mbox{with} \quad
g_1 = \frac{cos(\theta)(x-t_x) + sin(\theta) (y-t_y)}{s_x} \\ \nonumber
\mbox{and} \quad
g_2= \frac{cos(\theta)(y-t_y) - sin(\theta) (x-t_x)}{s_y}.
\end{eqnarray}
In Fig.~\ref{Fig:gaussian_atoms} we illustrate Gaussian atoms for different translation, rotation and anisotropic scaling parameters. Now, we can write the linear approximation of the reference image $\hat{I}_1$ with functions in $\mathcal{D}$ as
\begin{equation} \label{eqn:I1}
 \hat{I}_1 \approx \sum_{k=1}^{K} c_k~g_{\gamma_k},
\end{equation}
where $\{c_k\}$ represents the coefficient vector. The $K$ number of atoms used in the approximation of $\hat{I}_1$ is usually much smaller than the dimension of the image $\hat{I}_1$. We use here a suboptimal solution based on matching pursuit \cite{Mallat,Figueras_TIP} in order to estimate the set of $K$ atoms. 

\begin{figure}
 \centering
$\begin{array}{@{\hspace{0.1in}} c@{\hspace{0.2in}}c} \multicolumn{1}{l}{\mbox{}} &  \multicolumn{1}{l}{\mbox{}}   \\
   \epsfxsize=1.5in \epsffile{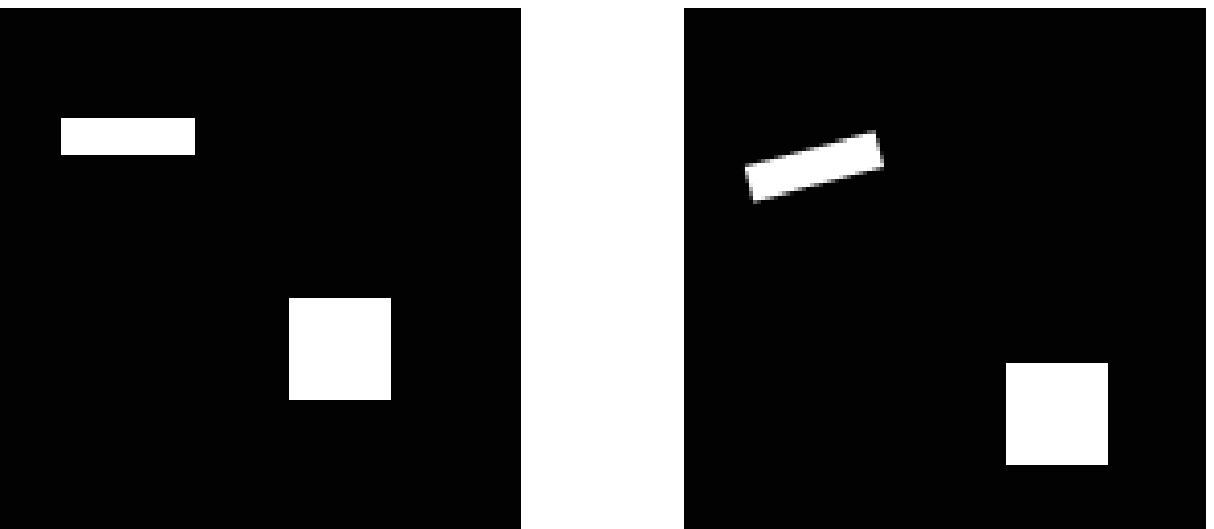} & \epsfxsize=1.5in \epsffile{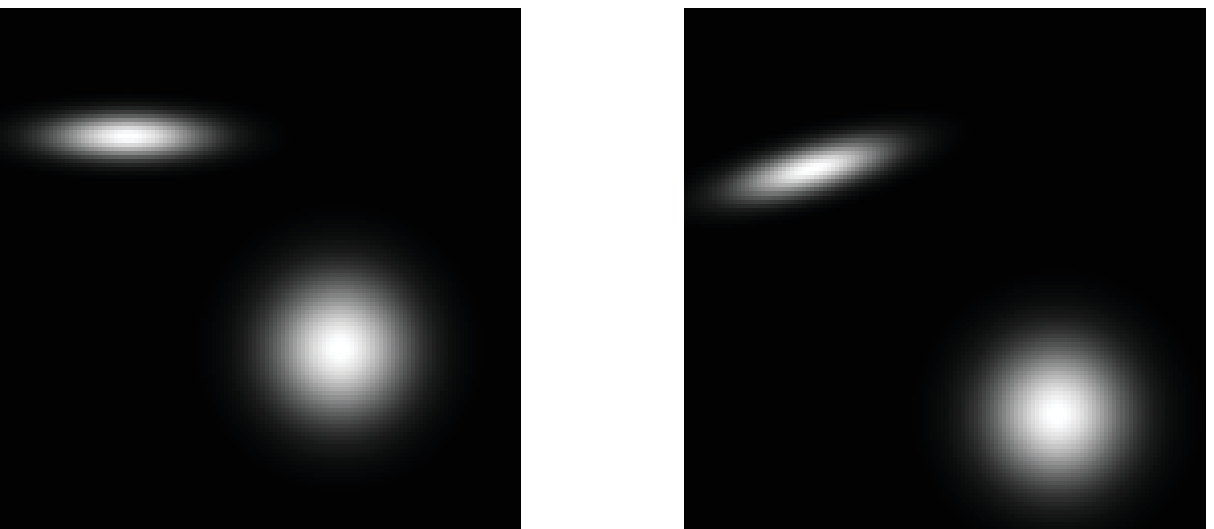}  \\      \mbox{(a) } & \mbox{(b) }   
   \end{array}$
\caption{Illustration of the atom transform $F$ in the approximation of the correlated images: (a) original correlated synthetic images; (b) sparse approximation of the images using atoms in the dictionary. The \emph{rectangle} and \emph{square} objects are related with transformations $F^1$ and $F^2$ respectively. } 
  \label{Fig:transform_illus}
  \end{figure}

The correlation between images can now be described by the geometric deformation of atoms in different images \cite{Ivana_TIP,Oscar_TIP}. Once the reference image $\hat{I}_1$ is approximated as given in Eq.~(\ref{eqn:I1}), the second image $I_2$ could be approximated with transformed versions of the atoms used in the approximation of $\hat{I}_1$. We can thus approximate $I_2$ as
\begin{equation} \label{eqn:I2}
 I_2 \approx \sum_{k=1}^{K} c_k~F^{k}(g_{{\gamma}_k}) = \sum_{k=1}^{K} c_k~g_{{\gamma}_k^\prime}, 
 \end{equation}
where $F^{k}(g_{\gamma_k})$ represents a local geometrical transformation of the atom $g_{\gamma_k}$. Due to the parametric form of the dictionary it is interesting to note that the transformation $F^{k}$ on $g_{\gamma_k}$ boils down to a transformation $\delta\gamma$ of the atom parameters, i.e.,
\begin{equation} \label{eqn:atom_transformation}
F^{k}(g_{\gamma_k}) = U(\delta\gamma)g_{\gamma_k} =
U( \delta\gamma \circ \gamma_k)g = g_{ \delta\gamma \circ {\gamma_k}}
= g_{\gamma_k^{\prime}}.
\end{equation}
For clarity, we show in Fig.~\ref{Fig:transform_illus} a sample synthetic correlated image pair and their sparse approximations using atoms in the dictionary. We see that the sparse approximations of images can be described with the transform $F$ of atom parameters.

The true transformations $\{F^{k}\}$ however are unknown in practical distributed coding applications. Therefore, the main challenge in our framework consists in estimating the local geometrical transformations $\{F^{k}\}$ when the second image $I_2$ is only available in the form of quantized linear measurements $\hat{y}_2$. 
%

\section{Correlation estimation from compressed linear measurements}
\label{sec:estimating_motion}

\subsection{Regularized optimization problem} \label{sec:cost_function}

We describe now our optimization framework for estimating the correlation between images. Given the set of $K$ atoms $\{g_{\gamma_k}\}$ that approximate the first image $\hat{I}_1$, the correlation estimation problem consists in finding the corresponding visual patterns in the second image $I_2$ that is given only by compressed random measurements $\hat{y}_2$. This is equivalent to finding the correlation between images $I_1$ and $I_2$ with the joint sparsity model based on local geometrical transformations, as described in Section \ref{sec:framework}. 

In more details, we are looking for a set of $K$ atoms in $I_2$ that correspond to the $K$ visual features $\{g_{\gamma_k}\}$ selected in the first image. We denote their parameters by $\Lambda$ where $\Lambda$ = $(\gamma^\prime_1,\gamma^\prime_2,\ldots,\gamma^\prime_K)$ for some $\gamma^\prime_k$, $\forall k, 1 \leq k \leq K$. We propose to select this set of atoms $\{g_{\gamma_k^\prime}\}$ in a regularized energy minimization framework as a trade-off between efficient approximation of $I_2$ and smoothness or consistency of the local transformations between images. The energy model $E$ proposed in our scheme is expressed as 
\begin{equation} \label{eqn:energy}\tag{{OPT-1}}
E(\Lambda) = E_d(\Lambda) + \alpha_1 E_s(\Lambda),
\end{equation}
where $E_d$ and $E_s$ represent the data and smoothness terms, respectively. The regularization parameter $\alpha_1$ balances the importance of the data and smoothness terms. The solution to our correlation estimation is given by the set of $K$ atom parameters $\Lambda^*$ that minimizes the energy $E$, i.e.,
\begin{equation} \label{eqn:Opt}
\Lambda^* = arg \min_{\Lambda \in S}{E(\Lambda)}.
\end{equation}
The parameter $S$ represents the search space given by 
\begin{equation}\label{eqn:search_space}
S=   \left\{ (\gamma^\prime_1,\gamma^\prime_2,\ldots,\gamma^\prime_K)~ | ~ \gamma_k^{\prime} =  \delta\gamma \circ \gamma_k, 
       1 \leq k \leq K, \delta\gamma \in  \mathcal{L} \right\}.
\end{equation}
The multidimensional search window $\mathcal{L} \subset \mathbb{R}^5$ is defined as $\mathcal{L} = [-\delta t_x,~\delta t_x] \times [-\delta t_y,~\delta t_y] \times [-\delta \theta,~\delta \theta] \times [-\delta s_x,~\delta s_x] \times [-\delta s_y,~\delta s_y]$
where $\delta t_x, \delta t_y, \delta \theta, \delta s_x,\delta s_y $ determine the window size for each of the atom parameters (i.e., translations $t_x,t_y$, rotation $\theta$ and scales $s_x,s_y$). Even if our formulation is able to handle complex transformations, they generally take the form of motion vectors or disparity information in video coding or stereo imaging applications. The label sets and the search space $S$ are drastically reduced in this case. The terms used in OPT-1 are described in the next paragraphs. 

\subsection{Data cost function} \label{sec:data_cost}
The data cost function computes (in the compressed domain) the accuracy of the sparse approximation of the second image with geometric atoms linked to the reference image. The decoder receives the measurements $\hat{y}_2$ that are computed by the quantized projections of $I_2$ onto a sensing matrix $\Phi$. For each set of $K$ atom parameters $\Lambda = \{\gamma^\prime_k\}$ the data term $E_d$ reports the error between measurements $\hat{y}_2$ and orthogonal projection of $\hat{y}_2$ onto $\Psi_\Lambda$ that is formed by the compressed versions of the atoms, i.e., $\Psi_\Lambda  = \Phi[g_{\gamma^\prime_1}|g_{\gamma^\prime_2}|\ldots|g_{\gamma^\prime_K}]$. It turns out that the orthogonal projection of $\hat{y}_2$ is given as $\Psi_{\Lambda} \Psi_{\Lambda}^{\dagger}\hat{y}_2$, where $\dagger$ represents the pseudo-inverse operator. More formally, the data cost is computed using the following relation:
\begin{equation} \label{eqn:datacost}
E_d(\Lambda) = \norm{\hat{y}_2 - \Psi_{\Lambda}
\Psi_{\Lambda}^{\dagger} \hat{y}_2}_2^2= \norm{\hat{y}_2 - \Psi_{\Lambda}
c}_2^2.
\end{equation}
The data cost function given in Eq. ~(\ref{eqn:datacost}) therefore first calculates the coefficients $c = \Psi_\Lambda^\dagger \hat{y}_2$ and then measures the distance between the observations $\hat{y}_2$ and $\Psi_\Lambda c$. In other words, the data cost function $E_d$ accounts for the intensity variations between images by estimating the coefficients $c$ of the warped atoms.

When the measurements are quantized, the coefficient vector $c$ fails to properly account for the error introduced by the quantization. The quantized measurements only provide the index of the quantization interval containing the actual measurement value and the actual measurement value could be any point in the quantization interval. Let $y_{2}(i)$ be the $i^{th}$ coordinate of the original measurement and $\hat{y}_{2}(i)$ be the corresponding quantized value. It can be noted that the joint decoder has only access to the quantized value $\hat{y}_{2}(i)$ and not the original value $y_{2}(i)$. Henceforth, the joint decoder knows that the quantized measurement lies within the quantization interval, i.e., $ \hat{y}_{2}(i) \in \mathcal{R}_{\hat{y}(i)} = (r_i ~ r_{i+1}]$ where $r_i$ and $r_{i+1}$ define the lower and upper bounds of the quantizer bin $\mathcal{Q}_i$. We therefore propose to refine the data term in the presence of quantization by computing a coefficient vector $\tilde{c}$ as the most consistent coefficient vector when considering all the possible measurement vectors that can result in the quantized measurements vector $\hat{y}_2$. In more details, the quantized measurements $\hat{y}_2$ can be produced by all the observation vectors $\tilde{y}_2 \in \mathcal{R}_{\hat{y}} $, where $\mathcal{R}_{\hat{y}}$ is the Cartesian product of all the quantized regions $ \mathcal{R}_{\hat{y}(i)}$, i.e., $\mathcal{R}_{\hat{y}}= \prod_{i}\mathcal{R}_{\hat{y}(i)}$. The data cost term given in Eq.~(\ref{eqn:datacost}) can thus be modified as
\begin{equation} \label{eqn:robustdatacost}
\tilde{E}_d(\Lambda)  = \min_{\tilde{c},\tilde{y}_2} \norm{\tilde{y}_2 - \Psi_\Lambda \tilde{c}}_2^2,  ~ \mbox{s.t.} ~ \tilde{y}_2  \in \mathcal{R}_{\hat{y}}.
\end{equation}
Therefore, the robust data term $\tilde{E}_d(\Lambda)$ first jointly estimates the coefficients $\tilde{c}$ and the measurements $\tilde{y}_2$, and then computes the distance between the $\tilde{y}_2$ and $\Psi_\Lambda \tilde{c}$. It can be shown that the Hessian of the objective function $h(\tilde{c},\tilde{y}_2) = \parallel \tilde{y}_2 - \Psi_\Lambda \tilde{c} \parallel_2^2$ in Eq.~(\ref{eqn:robustdatacost}) is positive semidefinite, i.e., $\nabla^2h \succeq 0 $, and hence the objective function $h$ is convex. Also, the region $\mathcal{R}_{\hat{y}}$ forms a closed convex set as each region $\mathcal{R}_{\hat{y}_i}= (r_i ~ r_{i+1}]$, $\forall i$ forms a convex set. Henceforth, the optimization problem given in Eq.~(\ref{eqn:robustdatacost}) is convex, which leads to effective solutions.

\subsection{Smoothness cost function} \label{sec:smoothness_cost}

The goal of the smoothness term $E_s$ in OPT-1 is to regularize the atom transformations such that the transformations are coherent for neighbor atoms. In other words, the atoms in a spatial neighborhood are likely to undergo similar transformations, when the correlation between images is due to object or camera motion. Instead of penalizing directly the transformation $\{F^k\}$ to be coherent for neighbor atoms, we propose to generate a dense disparity (or motion) field from the atom transformations and to penalize the disparity (or motion) field such that it is coherent for adjacent pixels. This regularization is easier to handle than a regular set of transformations $\{F^k\}$ and directly corresponds to the physical constraints that explain the formation of correlated images.  

In more details, for a given transformation value ${\delta \gamma = (t_x^\prime-t_x, t_y^\prime-t_y, \theta^\prime-\theta, s_x/s_x^\prime,s_y/s_y^\prime)}$ at pixel $\bold{z}$ we compute the horizontal component $\bold{m}^h$ and vertical component $\bold{m}^v$ of the motion field as 
\begin{eqnarray} \label{eqn:dense_gt}
\left[ \begin{array}{c} \bold{m}^h(\bold{z})\\   \bold{m}^v(\bold{z}) \end{array} \right]  & = &
\left[ \begin{array}{c} m(\bold{z})-t_x \\  n(\bold{z})-t_y \end{array} \right] -
{S} {R}{T}  
\end{eqnarray}
where $(m(\bold{z}), n(\bold{z}))$ represent the Euclidean coordinates. The matrices $S$, $R$ and $T$ represent the grid transformations due to scale, rotation and translation changes respectively. They are defined as 
\begin{equation} \nonumber
 S  =  \begin{bmatrix}
  s_x/s_x^\prime & 0\\ 
0 & s_y/s_y^\prime
 \end{bmatrix},
 R =   \begin{bmatrix}
  cos(\theta^\prime-\theta) & sin(\theta^\prime-\theta)\\
 -sin(\theta^\prime-\theta) & cos(\theta^\prime-\theta)
 \end{bmatrix}, \\
 \end{equation}
and
\begin{equation} \nonumber
T  =  \left[ \begin{array}{c} m(\bold{z})-t_x- (t_x^\prime-t_x) \\  n(\bold{z})-t_y-(t_x^\prime-t_x) \end{array} \right].
\end{equation}
Finally, the smoothness cost $E_s$ in OPT-1 is given as
\begin{equation}\label{eqn:smooth_term}
E_s(\Lambda)  = \sum_{\bold{z},\bold{z^\prime} \in \mathcal{N}}
V_{\bold{z},\bold{z^\prime}},
\end{equation}
where $\bold{z},\bold{z^\prime}$ are the adjacent pixel locations and $\mathcal{N}$ is the usual 4-pixel neighborhood. The term $V_{\bold{z},\bold{z^\prime}}$ in Eq.~(\ref{eqn:smooth_term}) captures the distance between local transformations in neighboring pixels. It is defined as 
\begin{equation} \label{eqn:Vterm}
V_{\bold{z},\bold{z^\prime}} =  \min \left(|\bold{m}^h(\bold{z}) -\bold{m}^h(\bold{z^{\prime}})| + |\bold{m}^v(\bold{z}) -\bold{m}^v(\bold{z^{\prime}})|, \tau \right). 
\end{equation}
 The parameter $\tau$ in Eq.~(\ref{eqn:Vterm}) sets a maximum limit to the penalty; it helps to preserve the discontinuities in the transformation field that exist at boundaries of visual objects \cite{Graph_cuts}. 


\subsection{Optimization algorithm}\label{sec:opt_algo}

We describe now the optimization methodology that is used solve OPT-1. Recall that our objective is to assign a transformation $F$ to each atom $g_{\gamma_k}$ in the reference image in order to build a set of smooth local transformations that is consistent with the quantized measurements $\hat{y}_2$. The candidate transformations are chosen from a finite set  of labels $\mathcal{L} = \mathcal{L}_x \times \mathcal{L}_y \times \mathcal{L}_{\theta} \times \mathcal{L}_a\times \mathcal{L}_b$ where $\mathcal{L}_{x}$, $\mathcal{L}_{y}$, $\mathcal{L}_{\theta}$, $\mathcal{L}_{a}$ and $\mathcal{L}_{b}$ refer to the label sets corresponding to translation along $x$ and $y$ directions, rotations and anisotropic scales respectively (see Eq.~(\ref{eqn:search_space})). One could use an exhaustive search on the entire label $\mathcal{L}$ to solve OPT-1. However, the cost for such a solution is high as the size of the label set $\mathcal{L}$ grows exponentially with the size of the search windows $\delta t_x, \delta t_y, \delta \theta, \delta s_x,\delta s_y $. Rather than doing an exhaustive search, we use graph-based minimization techniques that converge to strong local minima or global minima in a polynomial time with tractable computational complexity \cite{Graph_cuts,GC_comp_anal}.

\begin{figure}[!t]
\centering
\epsfxsize=3in \epsffile{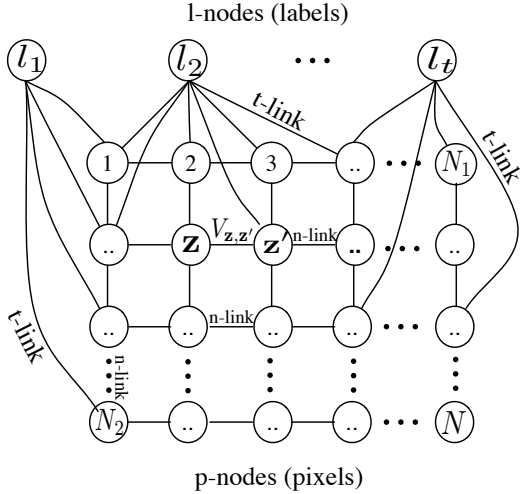}
 \caption{ A graph $\mathcal{G} = (\mathcal{V},\mathcal{E})$ is constructed using the set of vertices $\mathcal{V} = \mathcal{Z} \cup \mathcal{L}$, where the pixels nodes $\mathcal{Z} = \{1,2,\ldots,N \}$ and label nodes $\mathcal{L} = \{l_1,l_2,\ldots,l_t \}$. Each pixel $\bold{z}$ is connected to the l-node with a t-link. Some t-links are omitted for the sake of clarity. The pixels $\bold{z}, \bold{z}^\prime \in \mathcal{N}$ are connected with a n-link. The correlation solution is given a multiway cut that leaves each p-node connected with only one t-link \cite{Graph_cuts}.} 
  \label{Fig:gc_illus}
  \end{figure}


Usually in Graph Cut algorithms a graph $\mathcal{G} = (\mathcal{V},\mathcal{E})$ is constructed using set of vertices $\mathcal{V}$ and edges $\mathcal{E}$. The set of vertices are given as $\mathcal{V} = \mathcal{Z} \cup \mathcal{L}$, where $\mathcal{Z}$ define of nodes corresponding to the pixels in the images (p-nodes) and $\mathcal{L}$ define the label nodes (l-nodes), as shown in Fig.~\ref{Fig:gc_illus}. The p-nodes that are in the neighborhood $\mathcal{N}$ are connected by an edge called n-link. The cost of n-link usually corresponds to the penalty of assigning different labels to the adjacent pixels as given by $V_{\bold{z},\bold{z}^\prime}$. Also, each p-vertex $\bold{z}$ is connected to the l-node by an edge called t-link. The cost of a t-link connecting a pixel and a label corresponds to the penalty of assigning the corresponding label to that pixel; this cost is normally derived from the data term. The final solution is given by a multi-way cut that leaves each p-vertex connected with exactly one t-link. For more details we refer the reader to \cite{Graph_cuts}. 

In order to solve our OPT-1 problem, we first need to map our cost functions on the graph in order to assign weights to the n-links and t-links. For a given pair of transformation labels at pixels $\bold{z}$ and $\bold{z}^\prime$, it is straightforward to calculate the weights of the n-links using Eq.~(\ref{eqn:Vterm}). It should be noted that the motion field for a given label is computed using Eq.~(\ref{eqn:dense_gt}). We now describe how to calculate the cost of the t-links based on the data cost $E_d(\Lambda)$. 
Let $\mathcal{Z}_k$ be the set of pixels in the support of the atom $g_{\gamma_k}$ that is given as  
\begin{equation}\label{eqn:atom_support}
\mathcal{Z}_k = \{  \bold{z} = (x,y) | g_{\gamma_k}(x,y) > \epsilon \},
\end{equation}
where $\epsilon > 0$ is a constant. Using this definition, we calculate the t-link penalty cost of connecting a label node $l_k \in \mathcal{L}$ to all the pixel nodes $\bold{z}$ in the support of the atom $g_{\gamma_k}$ as $E_d(\Lambda)$ given in Eq.~(\ref{eqn:datacost}), where $\Lambda = {(\gamma_1,\gamma_2,\ldots, l_k \circ \gamma_k,\ldots,\gamma_K)}$. That is, the t-link cost computed between the label $l_k$ and pixels $\bold{z}, \forall \bold{z} \in \mathcal{Z}_k$ is $E_d(\Lambda)$. However, due to atom overlapping the pixels in the overlapping region could be assigned more than one label. In such cases, we compute the cost corresponding to the index $k^\prime$ of the atom that has the maximum atom response. The index $k^\prime$ is computed as 
\begin{equation} \label{eqn:smth_kprime}
k^\prime = \displaystyle arg \max_{k=1,2,\ldots,K}w^{(k)}_\bold{z}, 
\end{equation}
where $w^{(k)}_\bold{z}$ is the response of the  $k^{th}$ atom  at the location $\bold{z}$, i.e., $w^{(k)}_\bold{z}=g_{\gamma_k}(\bold{z})=g_{\gamma_k}(x,y)$. After mapping the cost functions on the graph we calculate the correlation solution using a max-flow/min-cut algorithm \cite{Graph_cuts}. Finally, the data term $E_d$ in OPT-1 can be replaced with the robust data term $\tilde{E}_d$ given in Eq.~(\ref{eqn:robustdatacost}) in order to provide robustness to quantization errors. The resulting optimization problem can be efficiently solved using Graph Cut algorithms as described above. 

\subsection{Complexity considerations}

We discuss now briefly the computational complexity of our correlation estimation algorithm which can basically be divided into two stages. The first stage finds the most prominent features in the reference image using sparse approximations in a structured dictionary. The second stage estimates the transformation for all the features in the reference image by solving a regularized optimization problem  OPT-1.

Overall, our framework offers a very simple encoding stage with image acquisition based on random linear projections. The computational burden is shifted to the joint decoder which can still trade-off complexity and performance.  Even if the decoder is able to handle computationally complex tasks in our framework, the complexity of our system stays reasonable due to the efficiency of Graph Cuts algorithms whose complexity is bounded by a low order polynomial \cite{Graph_cuts,GC_comp_anal}. The complexity can be further reduced in both stages compared to the generic implementation proposed above. For example, the complexity of the sparse approximation of the reference image can be reduced significantly using a tree-structured dictionary, without significant loss in the approximation performance \cite{Jost_TreeMP}. In addition, a block-based dictionary can be used in order to reduce the complexity of the transformation estimation problem with block-based computations. Experiments show however that this comes at a price of a performance penalty in the reconstruction quality. Overall, it is clear that the decoding scheme proposed above offers high flexibility with an interesting trade-off between the complexity and the performance. For example, one might decide to use the simple data cost $E_d$ even when the measurements are quantized; it leads to a simpler scheme but to a reduced reconstruction quality.

\section{Consistent image prediction by warping} \label{sec:cons_pred}
After correlation estimation, one can simply reconstruct an approximate version of the second image $\hat{I}_2$ by warping the reference image $\hat{I}_1$ using a set of local transformations that forms the warping operator $\mathcal{W}_{\Lambda}$ (see Fig.~\ref{Fig:block_scheme}). The resulting approximation is however not necessarily consistent with the quantized measurements $\hat{y}_2$; the measurements corresponding to the projection of the image $\hat{I}_2$ on the sensing matrix $\Phi$ are not necessarily equal to $\hat{y}_2$. The consistency error might be significant, because the atoms used to compute the correlation and the warping operator do not optimally handle the texture information. 

We therefore propose to add a consistency term $E_t$ in the energy model of OPT-1 and to form a new optimization problem for improved image prediction. The consistency term forces the image reconstruction through the warping operator to be consistent with the quantized measurements. We define this additional term $E_t$ as the $l_2$ norm error between the quantized measurements generated from the reconstructed image $\hat{I}_2= \mathcal{W}_{\Lambda}(\hat{I}_1)$ and the measurements $\hat{y}_2$. The consistency term $E_t$ is written as
\begin{equation}\label{eq:consistency}
E_t  (\Lambda)= \norm{\hat{y}_2 - \mathcal{Q}[\Phi\hat{I}_2]}_ 2  = \norm{\hat{y}_2 - \mathcal{Q}[\Phi\mathcal{W}_{\Lambda}(\hat{I}_1)] }_2,
\end{equation}
where $\mathcal{Q}$ is the quantization operator. In the absence of quantization the consistency term simply reads as
\begin{equation}\label{eq:mod_consistency}
E_t (\Lambda)= \norm{y_2 - \Phi\mathcal{W}_{\Lambda}(\hat{I}_1)}_2.
\end{equation}
We then merge the three cost functions $E_d$, $E_s$ and $E_t$ with regularization constants $\alpha_1$ and $\alpha_2$ in order to form a new energy model $E_R$ for consistent reconstruction. It is given as
\begin{equation} \label{eqn:Opt_new} \tag{OPT-2}
E_R(\Lambda) = E_d(\Lambda) + \alpha_1 E_s(\Lambda) +  \alpha_2 E_t(\Lambda).
\end{equation}

We now highlight the differences between the terms $E_d$ and $E_t$ used in OPT-2. The data cost  $E_d$  adapts the coefficient vector to consider the intensity variations between images but it fails to properly handle the texture information. On the other hand, the consistency term $E_t$ warps the atoms by considering the texture information in the reconstructed image $\hat{I}_1$ but it fails to carefully deal with the intensity variations between images. These two terms therefore impose different constraints on the atom selection that effectively reduce the search space. We have observed experimentally that the quality of the predicted image $\hat{I}_2$ is maximized when all three terms are activated in the OPT-2 optimization problem.

 We propose to use the optimization method based on Graph Cuts described in Section~\ref{sec:opt_algo} in order to solve OPT-2. In particular, we map the consistency cost $E_t$ into the graph (see Fig.~\ref{Fig:gc_illus}) in addition to the data cost $E_d$ and smoothness cost $E_s$. For a given $\Lambda = {(\gamma_1,\gamma_2,\ldots, l_k \circ \gamma_k, \ldots, \gamma_K)}$, we propose to compute the t-link cost of connecting the label $l_k \in \mathcal{L}$ to the pixels $\bold{z}, \forall \bold{z} \in \mathcal{Z}_i$ as a cumulative sum of $E_d(\Lambda)+ \alpha_2 E_t(\Lambda)$. In the overlapping regions, as described earlier we take the value corresponding to the atom index $k^\prime$ that has maximum response as given in  Eq.~(\ref{eqn:smth_kprime}). Then, the n-link weights for the adjacent pixels $\bold{z}$ and $\bold{z}^\prime$ are computed based on Eq.~(\ref{eqn:Vterm}). After mapping the cost functions on the graph the correlation solution is finally estimated using max-flow/min-cut algorithms \cite{Graph_cuts}. Finally, the data cost $E_d$ in OPT-2 can be again replaced by the robust data term $\tilde{E}_d$ given in Eq.~(\ref{eqn:robustdatacost}). We show later that the performance of our scheme improves by using the robust data term $\tilde{E}_d$ in the presence of quantization.  At last,  the complexity of estimating the correlation model with OPT-2 problem is tractable, thanks to the efficiency of Graph Cuts algorithms \cite{Graph_cuts,GC_comp_anal}.

\section{Correlation estimation of multiple image sets} \label{sec:ext_mv}

So far, we have focused on the distributed representation of image pairs. In this section, we describe the extension of our framework to the datasets with $J$ correlated images denoted as $I_1,I_2, \ldots, I_J$. Similar to the stereo setup, we consider $I_1$ as the reference image. This image is given in a compressed form $\hat{I}_1$ and its prominent features are extracted at decoder with a sparse approximation over the dictionary $\mathcal{D}$ (see Section~\ref{sec:cost_function}). The images $I_2, \ldots, I_J$ are sensed independently using the measurement matrix $\Phi$ and their respective measurements $y_2, \ldots, y_J$ are quantized and entropy coded.  Our framework can be applied to image sequences or multi-view imaging. For the sake of clarity, we focus on a multi-view imaging framework where the multiple images are captured from different viewpoints. 

We are interested in estimating a depth map $Z$ that captures the correlation among $J$ images by assuming that the camera parameters are given a priori. The depth map is constructed using the set of $K$ features $\{g_{\gamma_k}\}$ in the reference image and the quantized measurements $\hat{y}_2, \ldots, \hat{y}_J$. We assume that the depth values $Z$ are discretized such that the inverse depth $1/Z$ is uniformly sampled in the range $[1/Z_{max},1/Z_{min}]$ where $Z_{min}$ and $Z_{max}$ are the minimal and maximal depth in the scene, respectively \cite{multiview_gc}. The problem is equivalent to finding a set of labels  $l \in \mathcal{L}$ that effectively captures the depth information for each atom $g_{\gamma_k}$ or pixel $\mathbf{z}$ in the reference image, where $\mathcal{L}$ is a discrete set of labels corresponding to different depths. We propose to estimate the depth information with an energy minimization problem OPT-3 which includes three cost functions as follows:
\begin{equation} \label{eqn:energy_multi-view} \tag{OPT-3}
H(\Lambda) = H_{d}(\Lambda) + \alpha_1 H_{s}(\Lambda) +  \alpha_2 H_{t}(\Lambda),
\end{equation}
where $H_{d}, H_{s} ~\mbox{and} H_{t}$  represent the data, smoothness and consistency terms respectively. These three terms are balanced with regularization constants $\alpha_1$ and $\alpha_2$. 

\begin{figure}[!t]
\centering
    \epsfxsize=3.2in \epsffile{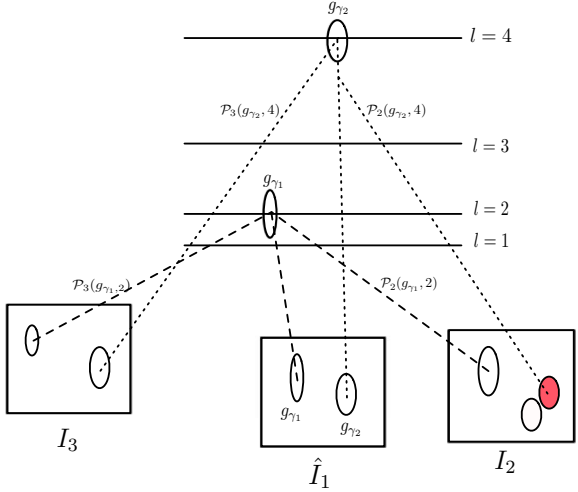}
 \caption{Illustration of the atom interactions in the multi-view imaging scenario. The original position of the features in all the images is marked in black color. Projection of the first feature $g_{\gamma_1}$ at $l = 2$ in the views $I_2$ and $I_3$ corresponds to the actual position of the feature in the respective views and thus forms a valid 3D region at $l = 2$. Meanwhile, the projection of the second  feature $g_{\gamma_2}$ at $l = 4$ corresponds to the actual position only in view $I_3$ but not in view $I_2$ (highlighted in red color). Hence, the second feature does not intersect at $l = 4$ which results in suboptimal solution at $l =4$. } 
  \label{Fig:atoms_illus}
  \end{figure}

The data term $H_{d}$ assigns a set of labels $l_1, l_2, \ldots, l_K$ respectively to the $K$ atoms $g_{\gamma_1},g_{\gamma_2}, \ldots, g_{\gamma_K}$ while respecting consistency with the quantized measurements. It reads as
\begin{equation}\label{eqn:data_multi-view}
H_{d}(\{l_k\}) = \sum_{j= 2}^J \norm{\hat{y}_j - \Psi_\Lambda ^j {\Psi_\Lambda ^j}^\dagger \hat{y}_j}^2_2, 
\end{equation}
where $\Psi_\Lambda ^j = \Phi [  \mathcal{P}_j(g_{\gamma_1},l_1),  \mathcal{P}_j(g_{\gamma_2},l_2), \ldots, \mathcal{P}_j(g_{\gamma_k},l_k),  \ldots, \\ \mathcal{P}_j(g_{\gamma_K},l_K) ]$. The operator $\mathcal{P}_j(g_{\gamma_k},l)$ represents the projection of the atom $g_{\gamma_k}$ to the $j^{th}$ view when the local transformation is given by the depth label $l$ (see Fig.~\ref{Fig:atoms_illus}). It can be noted that the data term in Eq.~(\ref{eqn:data_multi-view}) is similar to the data term described earlier for image pairs (see Eq.~(\ref{eqn:datacost})) except that the sum is computed for all the views. Depending on the relative position of the $j^{th}$ camera with respect to the reference camera, the projection $\mathcal{P}_j(g_{\gamma_k},l)$ can involve changes in the translation, rotation or scaling parameter, or combinations of them. Therefore, the projection $\mathcal{P}_j(g_{\gamma_k},l)$ of the atom $g_{\gamma_k}$ to the $j^{th}$ view approximately corresponds to another atom in the dictionary $\mathcal{D}$. It is interesting to note that the data cost is minimal if the projection of the atom $g_{\gamma_k}$ onto another view corresponds to its actual position in this view\footnote{we assume here that we have no occlusions.}. This happens when the depth label $l$ corresponds to the true distance to the visual object represented by the atom $g_{\gamma_k}$. For example, the projection of the feature $g_{\gamma_1}$ in Fig.~\ref{Fig:atoms_illus} corresponds to  the actual position of the  features in views $I_2$ and $I_3$. Therefore, the data cost for this feature $g_{\gamma_1}$ at label $l$=2 is minimal. On the other hand, the projection of the feature $g_{\gamma_2}$ is far from the actual position of the corresponding feature in the view $I_2$. The corresponding data cost $\norm{{{y}_2 - \Psi_\Lambda ^2 {\Psi_\Lambda ^2}^\dagger {y}_2}}_2^2$ is high in this case which indicates a suboptimal estimation of the depth label $l$. 


The smoothness cost $H_{s}$ enforces consistency in the depth label for the adjacent pixels $\bold{z}$ and $\bold{z^\prime}$. It is given as 
\begin{equation}\label{eqn:smooth_mutiview}
H_{s} = \sum_{\bold{z},\bold{z^\prime} \in \mathcal{N}}
\min \left(|  Z(\bold{z}) - Z(\bold{z^{\prime}})|, \tau \right),
\end{equation}
where $\tau$ is a constant and $\mathcal{N}$ represents the usual 4-pixel neighborhood. Finally, the consistency term $H_{t}$ favors depth labels that lead to image predictions that are consistent with the quantized measurements. We compute the consistency for all the views as the cumulative sum of terms $E_t$ given in Eq.~(\ref{eq:consistency}). More formally, the consistency term $H_{t}$ in the multi-view scenario is computed as 
\begin{eqnarray}\label{eq:consistency_multiview}
H_{t}(\{l_k\}) &= & \sum_{j =2}^J \norm{\hat{y}_j- \mathcal{Q}[\Phi\hat{I}_j]}_ 2    \\ \nonumber
&= & \sum_{j =2}^J \norm{\hat{y}_j - \mathcal{Q}[\Phi\mathcal{W}^j(\hat{I}_1, \{l_k\})]}_2,
\end{eqnarray}
where $\mathcal{W}^j(\hat{I}_1, \{l_k\})$ predicts the $j^{th}$ view using the set of labels $\{l_k\}$ and the set of $K$ atoms $\{g_{\gamma_k}\}$. Finally, the {OPT-3} optimization problem can be solved in polynomial time using the graph-based optimization methodologies described in Section \ref{sec:opt_algo}. In this case, the weights to the  t-links connecting between the label $l_k$ and the pixels $\bold{z},\forall \bold{z} \in \mathcal{Z}_k$ are assigned as $H_d+ \alpha_2 H_t$. The n-link cost for the neighboring pixels $\bold{z}, \bold{z}^\prime \in \mathcal{N}$  is assigned as $\min \left(|  Z(\bold{z}) - Z(\bold{z^{\prime}})|, \tau \right)$.


\section{Experimental results}
\label{sec:Results}

\begin{figure*}
  \begin{center}
   $\begin{array}{@{\hspace{0.25in}} c@{\hspace{0.25 in}} c@{\hspace{0.25 in}} c@{\hspace{0.25 in}}c@{\hspace{0.25 in}}c} 
\multicolumn{1}{l}{\mbox{}} &\multicolumn{1}{l}{\mbox{}} &  \multicolumn{1}{l}{\mbox{}} &  \multicolumn{1}{l}{\mbox{}}&  \multicolumn{1}{l}{\mbox{}}\\
  \epsfxsize=1.1in {\epsffile{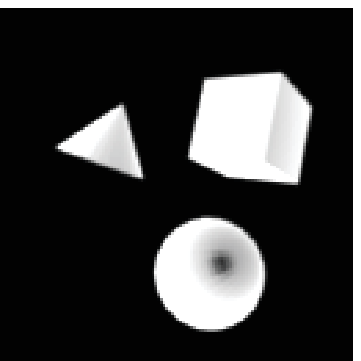}} & \epsfxsize=1.1in {\epsffile{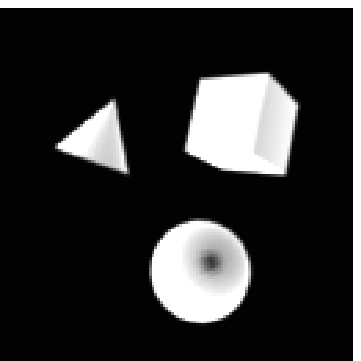}} &  \epsfxsize=1.1in {\epsffile{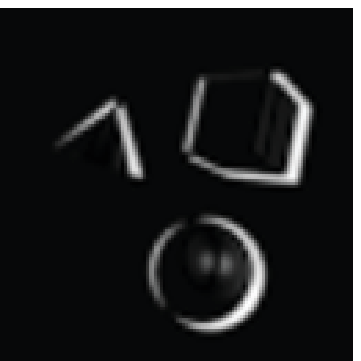}}  & \epsfxsize=1.1in {\epsffile{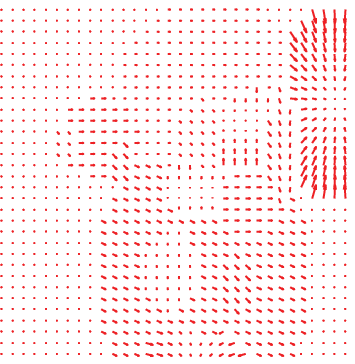}} & \epsfxsize=1.1in {\epsffile{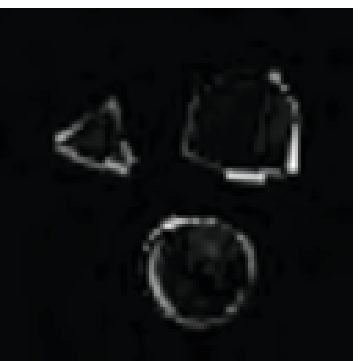}} \\ 
    \mbox{(a) $I_1$ } & \mbox{(b) $I_2$} & \mbox{(c) $|I_1-I_2|$ } & \mbox{(d) $(\bold{m}^h,\bold{m}^v)$ } & \mbox{(e) $|I_2-\hat{I}_2|$} \\ 
     \mbox{ } & \mbox{} & \mbox{} & \mbox{$E_s= 4309$ } & \mbox{} \\ \\
 \epsfxsize=1.1in {\epsffile{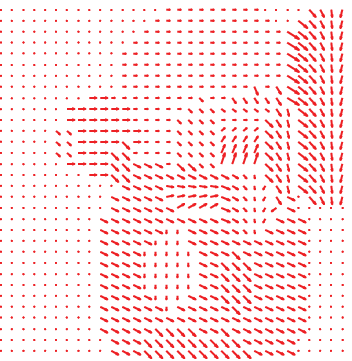}} &  \epsfxsize=1.1in {\epsffile{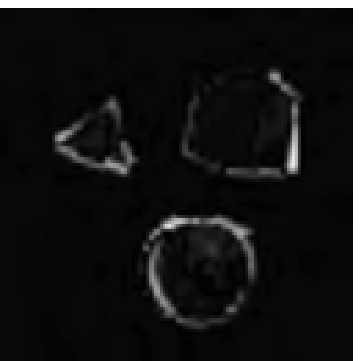}} & \epsfxsize=1.1in {\epsffile{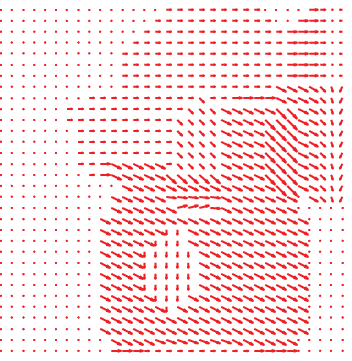}} &  \epsfxsize=1.1in {\epsffile{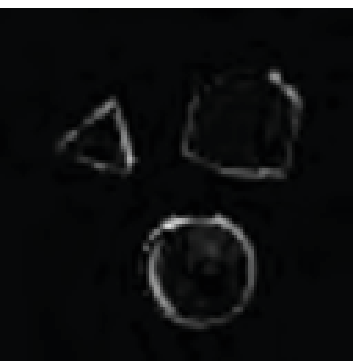}} \\ 
  \mbox{(f) $(\bold{m}^h,\bold{m}^v)$} & \mbox{(g) $|I_2-\hat{I}_2|$ } & \mbox{(h) $(\bold{m}^h,\bold{m}^v)$ } &  \mbox{(i) $|I_2-\hat{I}_2|$} \\
  \mbox{$E_s= 4851$} & \mbox{} & \mbox{$E_s= 1479$} &  \mbox{ } \\
     \end{array}$
  \end{center}  
\caption{Comparison of the estimated motion fields and the predicted images with the OPT-1 and OPT-2 problems in the synthetic scene. The motion field is estimated using a measurement rate of 5\% with a 2-bit quantization. (a) Original image $I_1$; (b) original image $I_2$; (c) absolute error between $I_1$ and $I_2$; (d) motion field estimated with OPT-1 without activating $E_s$, i.e., $\alpha_1$ = 0; (e) prediction error with OPT-1 when motion field in (d) is used for image prediction; (f) motion field  estimated with OPT-2 without activating $E_s$; (e) prediction error with OPT-2 when motion field in (f) is used for image prediction; (h) motion field estimated with OPT-2; (i) prediction error with OPT-2 when motion field in (h) is used for image prediction. The smoothness energy $E_s$ of the motion fields are (d) 4309 (f) 4851 and (h) 1479.  The PSNR of the predicted images $\hat{I}_2$ in (e), (g) and (i) w.r.t. $I_2$ are 20 dB, 20.4 dB and 21.53 dB respectively.  }
 \label{Fig:syn_scene}
 \end{figure*}

\subsection{Setup}

In this section, we report the performance of the correlation estimation algorithms in stereo and multi-view imaging frameworks.  In order to compute a sparse approximation of the reference image at decoder, we use a dictionary $\mathcal{D}$ that is constructed using two generating functions, as explained in \cite{Figueras_TIP}. The first one consists of 2D Gaussian functions in order to capture the low frequency components (see Fig.~\ref{Fig:gaussian_atoms}). The second function represents a Gaussian function in one direction and the second derivative of a Gaussian in the orthogonal direction in order to capture the edges. The discrete parameters of the functions in the dictionary are chosen as follows. The translation parameters $t_x$ and $t_y$ take any positive value and cover the full height $N_1$ and width $N_2$ of the image. Ten rotation parameters are used between 0 and $\pi$ with increments of $\pi/18$. Five scaling parameters are equi-distributed in the logarithmic scale from $1$ to $N_1/8$ vertically, and $1$ to $N_2/9.77$ horizontally. The image $I_2$ is captured by random linear projections using a scrambled block Hadamard transform with a block size of 8 \cite{Gan_Eusipco}. The measurements $y_2$ are quantized using a uniform quantizer and the bit rate is computed by encoding the quantized measurements using an arithmetic coder. Unless stated differently, the parameters $\alpha_1$ and $\alpha_2$ in the optimization problems are selected based on trial and error experiments such that the estimated transformation field maximizes the quality of the predicted image $\hat{I}_2$.

\subsection{Generic transformation}

We first study the performance of our scheme with a pair of synthetic images that contains three objects. The original images $I_1$ and $I_2$ are given in Fig.~\ref{Fig:syn_scene}(a) and  Fig.~\ref{Fig:syn_scene}(b) respectively. It is clear that the common objects in the images have different positions and scales. The absolute error between the original images is given in  Fig.~\ref{Fig:syn_scene}(c), where the PSNR between $I_1$ and $I_2$ is found to be 15.6 dB.

We encode the reference image $I_1$ to a quality of $35$dB and the number of features used for the approximation of $\hat{I}_1$ is set to $K=15$. The transformation field is estimated with $\delta t_x=\delta t_y=3$ pixels,  $\delta s_x=\delta s_y= 2$ samples and $\delta \theta = 0$. We first estimate the transformation field with the OPT-1 problem by setting $\alpha_1 = 0$, i.e., the smoothness term $E_s$ is not activated. The resulting motion field is shown in Fig.~\ref{Fig:syn_scene}(d). From Fig.~\ref{Fig:syn_scene}(d) we observe that the proposed scheme gives a good estimation of the transformation field even with a  $5\%$ measurement rate that are quantized with $2$ bits. We further see that the image $\hat{I}_2$ predicted with help of the estimated correlation information is closer to the original image $I_2$ than to $I_1$ (see Fig.~\ref{Fig:syn_scene}(e)). We then include the consistency term in addition to the data cost and we solve the problem OPT-2 without activating the smoothness term, i.e., $\alpha_1 = 0$. The estimated transformation field and the prediction error are shown in Fig.~\ref{Fig:syn_scene}(f) and Fig.~\ref{Fig:syn_scene}(g), respectively. We observe that the consistency term improves the quality of the motion field and the prediction quality. Finally, we highlight the benefit of enforcing smoothness constraint in our OPT-2 problem. The estimated transformation field with the OPT-2 problem including the smoothness term is shown in Fig.~\ref{Fig:syn_scene}(h). By comparing the motion fields in Fig.~\ref{Fig:syn_scene}(d) and Fig.~\ref{Fig:syn_scene}(f) we see that the motion field in Fig.~\ref{Fig:syn_scene}(h) is smoother and more coherent; this confirms the benefit of the smoothness term. Quantitatively, the smoothness energy $E_s$ of the motion field shown in Fig.~\ref{Fig:syn_scene}(h) is 1479, which is clearly smaller comparing to the solutions given Fig.~\ref{Fig:syn_scene}(d) and Fig.~\ref{Fig:syn_scene}(f) (resp. 4309 and 4851). Also, the smoothness term effectively improves the quality of the predicted image and the predicted image $\hat{I}_2$ gets closer to the original image $I_2$ as shown in Fig.~\ref{Fig:syn_scene}(i).

\subsection{Stereo image coding}
We now study the performance of our distributed image representation algorithms in stereo imaging frameworks. We use two datasets, namely \emph{Plastic} and \emph{Sawtooth}\footnote{These image sets are available at http://vision.middlebury.edu/stereo/data/ }. The images are downsampled to a resolution $N_1=144$, $N_2=176$ (original resolution of the datasets are $370\times423$ and $434\times380$ respectively). We carry out experiments using the views 1 and 3 for the Plastic dataset and views 1 and 5 for the Sawtooth dataset. These datasets have been captured by a camera array where the different viewpoints are uniformly arranged on a line. As this corresponds to translating the camera along one of the image coordinate axis, the disparity estimation problem becomes a one-dimensional search problem and the smoothness term in Eq.~(\ref{eqn:smooth_term}) is simplified accordingly. The viewpoint 1 is selected as the reference image $I_1$ and it is encoded such that the quality of $\hat{I}_1$ is approximately $33$~dB. Matching pursuit is then performed on $\hat{I}_1$ with $K = 30$ and $K = 60$ atoms for the Plastic and Sawtooth datasets respectively. The measurements on the second image are quantized using a 2-bit quantizer. At the decoder, the search for the geometric transformations $\{F^k\}$ is carried out along the translational component $t_x$ with window size $\delta t_x = 4$ pixels and no search is consider along the vertical direction, i.e., $\delta t_y = 0$. Unless stated explicitly, we use the data cost $E_d$ given in Eq.~(\ref{eqn:datacost}) in the OPT-1 and OPT-2  problems.   

\begin{figure*}[t!]
\centering
$\begin{array}{@{\hspace{-0.07in}} c@{\hspace{0.12 in}}c @{\hspace{0.12 in}} c@{\hspace{0.12 in}} c@{\hspace{0.12 in}} c} \multicolumn{1}{l}{\mbox{}} &  \multicolumn{1}{l}{\mbox{}} &\multicolumn{1}{l}{\mbox{}} &  \multicolumn{1}{l}{\mbox{}} &\multicolumn{1}{l}{\mbox{}} \\
   \epsfxsize=1.3in \epsffile{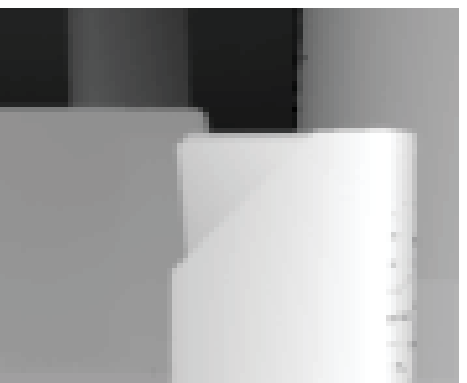} & \epsfxsize=1.3in \epsffile{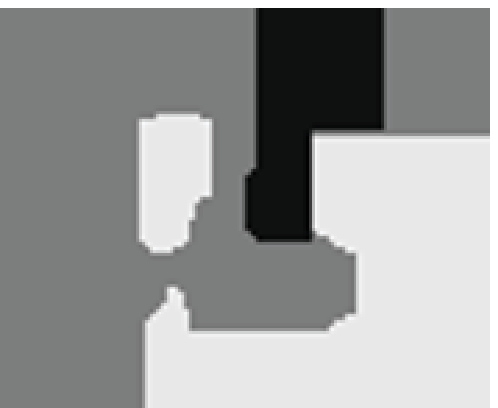} & \epsfxsize=1.3in \epsffile{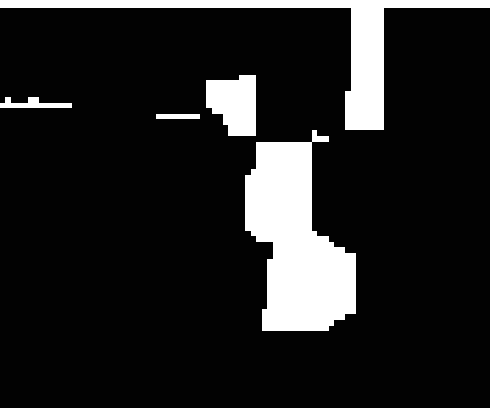} & \epsfxsize=1.3in \epsffile{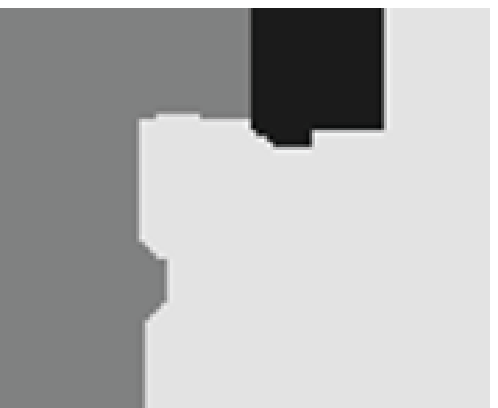} & \epsfxsize=1.3in \epsffile{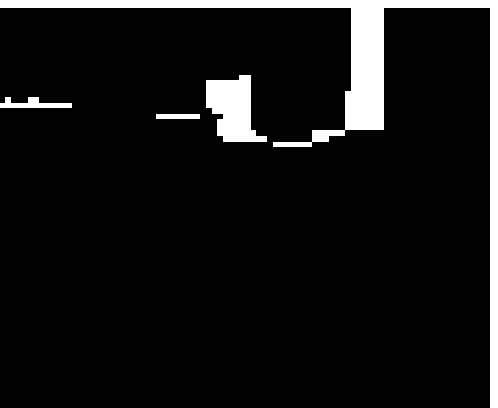} \\ 
     \mbox{(a) $\bold{M}^h$ } & \mbox{(b) $\bold{m}^h$ } & \mbox{(c) $ |\bold{M}^h - \bold{m}^h| > 1$ } & \mbox{(d) $\bold{m}^h$} & \mbox{(e) $ |\bold{M}^h - \bold{m}^h| > 1$}
  \end{array}$
\caption{Comparison of the estimated disparity fields with OPT-1 and OPT-2 for the Plastic dataset: (a) groundtruth disparity field $\bold{M}^h$ between views 1 and 2;  (b) estimated disparity field with OPT-1; (c) error in the disparity map with OPT-1 (DE = 10.8$\%$); (d)  estimated disparity field with OPT-2; (d) error in the disparity map with OPT-2 (DE = 4.1$\%$). The disparity field is estimated using a measurement rate of 35\% with a 2-bit quantization.}
  \label{Fig:plastic_depth_withoutreco}
  \end{figure*}
  
\begin{figure*}
\centering
 $\begin{array}{c@{\hspace{0.02 in}}c} \multicolumn{1}{l}{\mbox{}} &  \multicolumn{1}{l}{\mbox{}} \\
   \epsfxsize=3in \epsffile{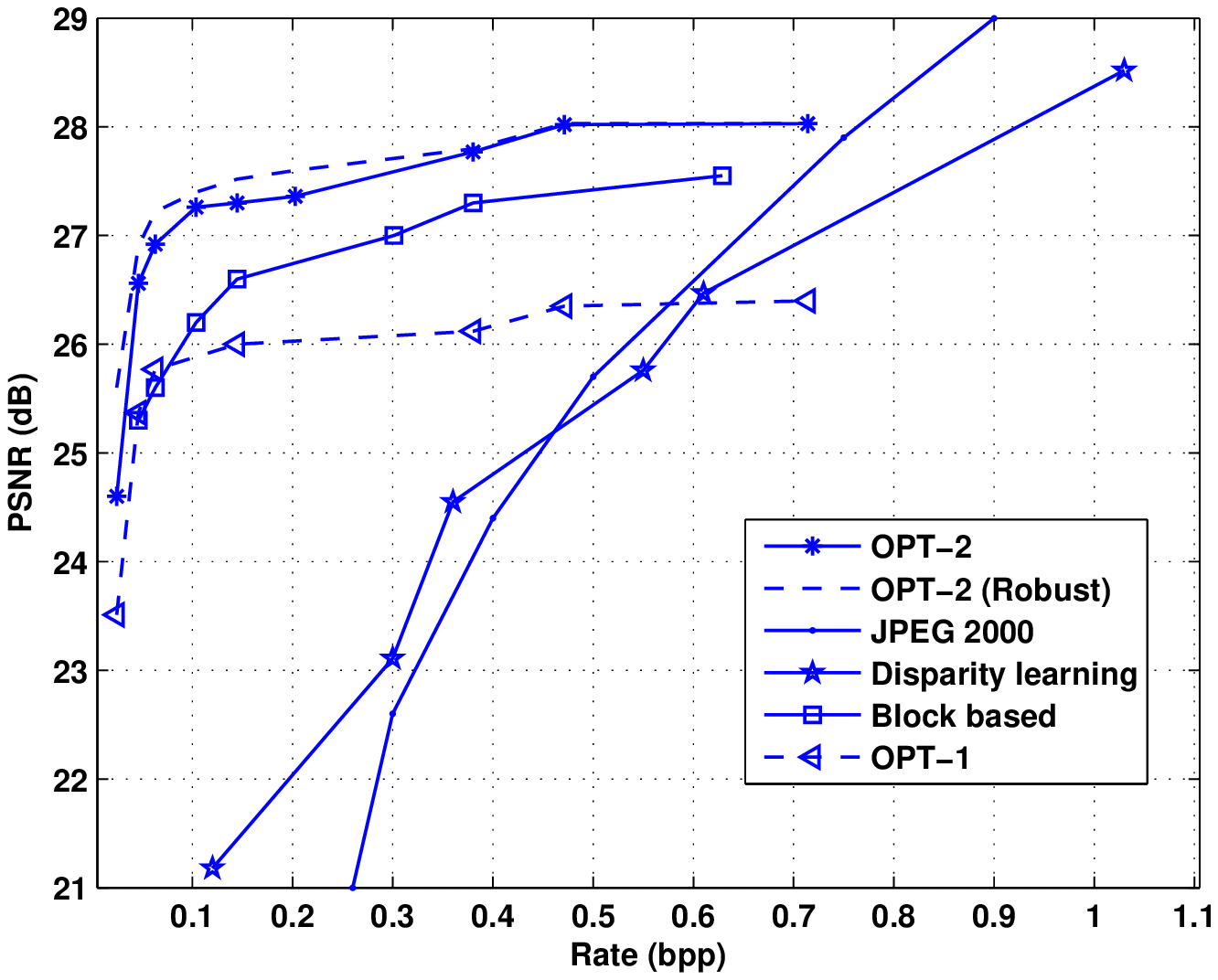} & \epsfxsize=3in \epsffile{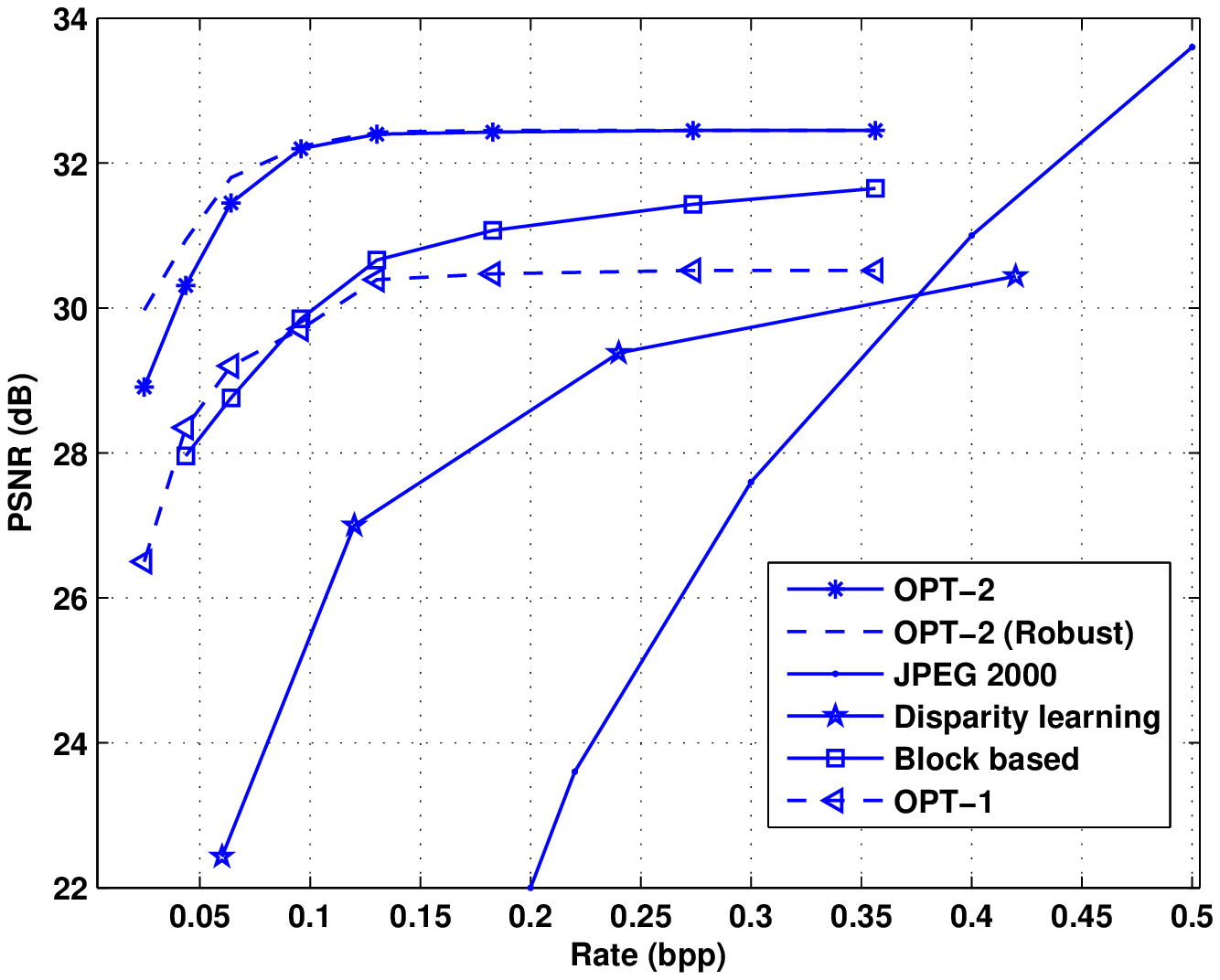} \\
   \mbox{(a) } & \mbox{(b)}
  \end{array}$
 \caption{Comparison of the RD performances between the proposed scheme, DSC scheme \cite{David}, block-based scheme \cite{YiMa_PCS} and independent coding solutions based on JPEG~2000 for (a) Sawtooth dataset, and (b) Plastic dataset.}
  \label{Fig:rdplot}
  \end{figure*}

We first study the accuracy of the estimated disparity information. In Fig.~\ref{Fig:plastic_depth_withoutreco}  we show the estimated disparity field $\bold{m}^h$ from $8870$ quantized measurements (i.e., a measurement rate of $35$\%) for the Plastic dataset. The groundtruth $\bold{M}^h$ is given in Fig.~\ref{Fig:plastic_depth_withoutreco}(a). The transformation is estimated by solving OPT-1 and the resulting dense disparity field is illustrated in Fig.~\ref{Fig:plastic_depth_withoutreco}(b). In this particular experiment, the parameter $\alpha_1$ is selected such that the error in the disparity map is minimized. The disparity error DE is computed between the estimated disparity field $\bold{m}^h$ and the groundtruth $\bold{M}^h$ as $DE = \frac{1}{N_1\times N_2} \sum_{\bold{z}=(x,y)} \left\{|\bold{M}^h(\bold{z}) - \bold{m}^h(\bold{z})| \geq 1\right\}$ where $N_1\times N_2$ represents the pixel resolution of the image \cite{Scharstein}. From Fig.~\ref{Fig:plastic_depth_withoutreco}(b) we observe that OPT-1 gives a good estimation of the disparity map; in particular the disparity value is correctly estimated in the regions with texture or depth discontinuities. We could also observe that the estimation of the disparity field is however less precise in smooth regions as expected from feature-based methods. Fortunately, the wrong estimation of the disparity value corresponding to the smooth region in the images does not significantly affect the warped or predicted image quality \cite{Scharstein}. Fig.~\ref{Fig:plastic_depth_withoutreco}(c) confirms such a distribution of the disparity estimation error where the white pixels denote an estimation error larger than one. We can see that the error in the disparity field is highly concentrated along the edges, since crisp discontinuities cannot be accurately captured due to the scale and smoothness of the atoms in the chosen dictionary. The disparity information estimated by OPT-2  is presented in Fig.~\ref{Fig:plastic_depth_withoutreco}(d) and the corresponding error is shown in Fig.~\ref{Fig:plastic_depth_withoutreco}(e). In this case, the regularization constants $\alpha_1$ and $\alpha_2$ in the OPT-2 problem are selected such that the DE is minimized. We see that the addition of the consistency term $E_t$ in the correlation estimation algorithm improves the performance.

We then study the rate-distortion (RD) performance of the proposed algorithms in the prediction of the image $\hat{I}_2$ in Fig. \ref{Fig:rdplot}. We show the performance of the reconstruction by warping the reference image according to the correlation computed by OPT-1 and OPT-2. We then highlight the benefit of using the robust data term $\tilde{E}_d$ in OPT-2 problem (denoted as \emph{OPT-2 (Robust)}). We use the optimization toolbox based on CVX~\cite{CVX_toolbox} in order to solve the optimization problem given in Eq.~(\ref{eqn:robustdatacost}). We then compare the RD performance to a distributed coding solution (DSC) based on the LDPC encoding of DCT coefficients, where the disparity field is estimated at the decoder using Expected Maximization (EM) principles \cite{David} (denoted as \emph{Disparity learning}). Then, in order to demonstrate the benefit of geometric dictionaries we propose a scheme denoted as \emph{block-based} that adaptively constructs the dictionary using blocks or patches in the reference image \cite{YiMa_PCS}. We construct a dictionary in the joint decoder from the reference image $\hat{I}_1$ segmented into $8\times8$ blocks. The search window size is $\delta t_x = 4$ pixels along the horizontal direction. We then use the optimization scheme described in OPT-2 to select the best block from the adaptive dictionary. In order to have a fair comparison, we encode the reference image $I_1$ similarly for both schemes (\emph{Disparity learning} and  \emph{block-based}) with a quality of $33$~dB (see Section \ref{sec:framework}). Finally, we also provide the performance of a standard JPEG~2000  independent encoding of the image $I_2$. From Fig.~\ref{Fig:rdplot}, we first see that the measurement consistency term $E_t$ significantly improves the decoding quality, as OPT-2 gives better performance than {OPT-1}. We further see that the OPT-2 problem with robust data cost improves the quality of the reconstructed image $\hat{I}_2$ by $0.5$-$1$~dB at low bit rates. Then, the results confirm that the proposed algorithms unsurprisingly outperform independent coding based on JPEG~2000, which outlines the benefits of the use of correlation in the decoding of compressed correlated images. At high rate, the performance of the proposed algorithms however tends to saturate as our model mostly handles the geometry and the correlation between images; but it is not able to efficiently handle the fine details or texture in the scene due to the image decoding $\hat{I}_2$ based on warping. From Fig.~\ref{Fig:rdplot}, it is then clear that the reconstruction of image $\hat{I}_2$ based on OPT-1 and OPT-2 outperforms the DSC coding scheme based on EM principles due to the accurate correlation estimation.  It is worth mentioning that state-of-the-art DSC scheme based on disparity learning compensate also for the prediction error in addition to correlation estimation. Though this is the case, our scheme outperforms DSC scheme due to an accurate disparity  field estimation. Finally, the experimental results also show that our schemes outperform the scheme based on block-based dictionary mainly because of the richer representation of the geometry and local transformations with the structured dictionaries.

\begin{figure}[!t]
\centering
    \epsfxsize=2.6in \epsffile{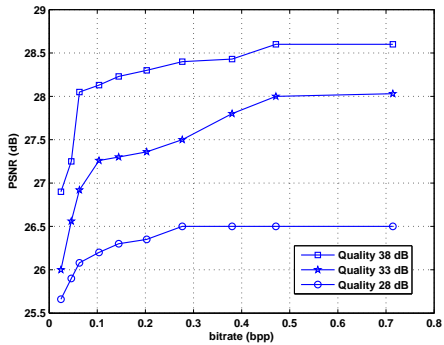}
 \caption{ RD performance with OPT-2 for decoding $\hat{I}_2$ (view 5) as a function of the quality of the reference image $\hat{I}_1$ (resp. 28 dB, 33 dB and 38 dB) in the Sawtooth dataset.}
  \label{Fig:rdplot_refquality}
  \end{figure}

\begin{figure}[t!]
\centering
    \epsfxsize=2.8in \epsffile{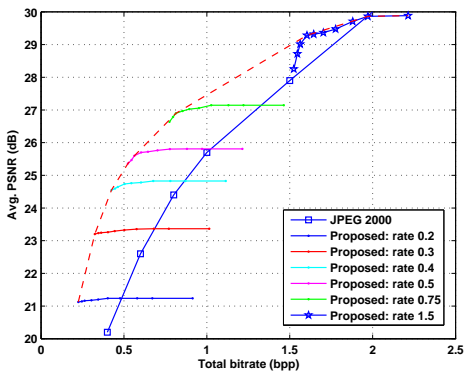}
 \caption{Overall RD performance between views 1 and 5 of the Sawtooth dataset. OPT-2 is used to predict the image $\hat{I}_2$ (view 5) using the image $\hat{I}_1$ (view 1) as the reference image. The image at view 5 is predicted with varying reference image bit rates $0.1, 0.2, 0.3, 0.4, 0.5, 0.75$ and $1.5$. }
  \label{Fig:rdplot_tbr}
  \end{figure}

We then study the influence of the quality of reference image $\hat{I}_1$ on the reconstruction performance. We use OPT-2 to reconstruct $\hat{I}_2$ (viewpoint 5) by warping when the reference image has been encoded at different qualities (i.e., different bit rates). Fig.~\ref{Fig:rdplot_refquality} shows that the reconstruction quality $\hat{I}_2$ improves with the quality of the reference image $\hat{I}_1$ as expected. While we have observed that the error in the disparity estimation is not dramatically reduced by improved reference quality, the warping stage permits to provide more details in the representation of $\hat{I}_2$ when the reference is of better quality. Finally, we study the overall RD performance for the Sawtooth dataset between views 1 and 5 that also includes the bit rate and quality of the reference image, in addition to the rate and quality of image $I_2$. Fig.~\ref{Fig:rdplot_tbr} shows the overall RD performance at reference image bit rates $0.2, 0.3, 0.4, 0.5, 0.75$ and $1.5$ bpp. In our experiments, for a given reference image quality we estimate the correlation model using OPT-2 (with 2-bit quantized measurements), and we compute the overall RD performance at that specific reference image bit rate. As shown before, the RD performance improves with increasing reference image quality. When we take the convex hull of the RD performances (which corresponds to implementing a proper rate allocation strategy), we outperform independent coding solutions based on JPEG~2000.

\begin{figure*}
\centering
 $\begin{array}{c@{\hspace{0.02 in}}c} \multicolumn{1}{l}{\mbox{}} &  \multicolumn{1}{l}{\mbox{}} \\
   \epsfxsize=2.8in \epsffile{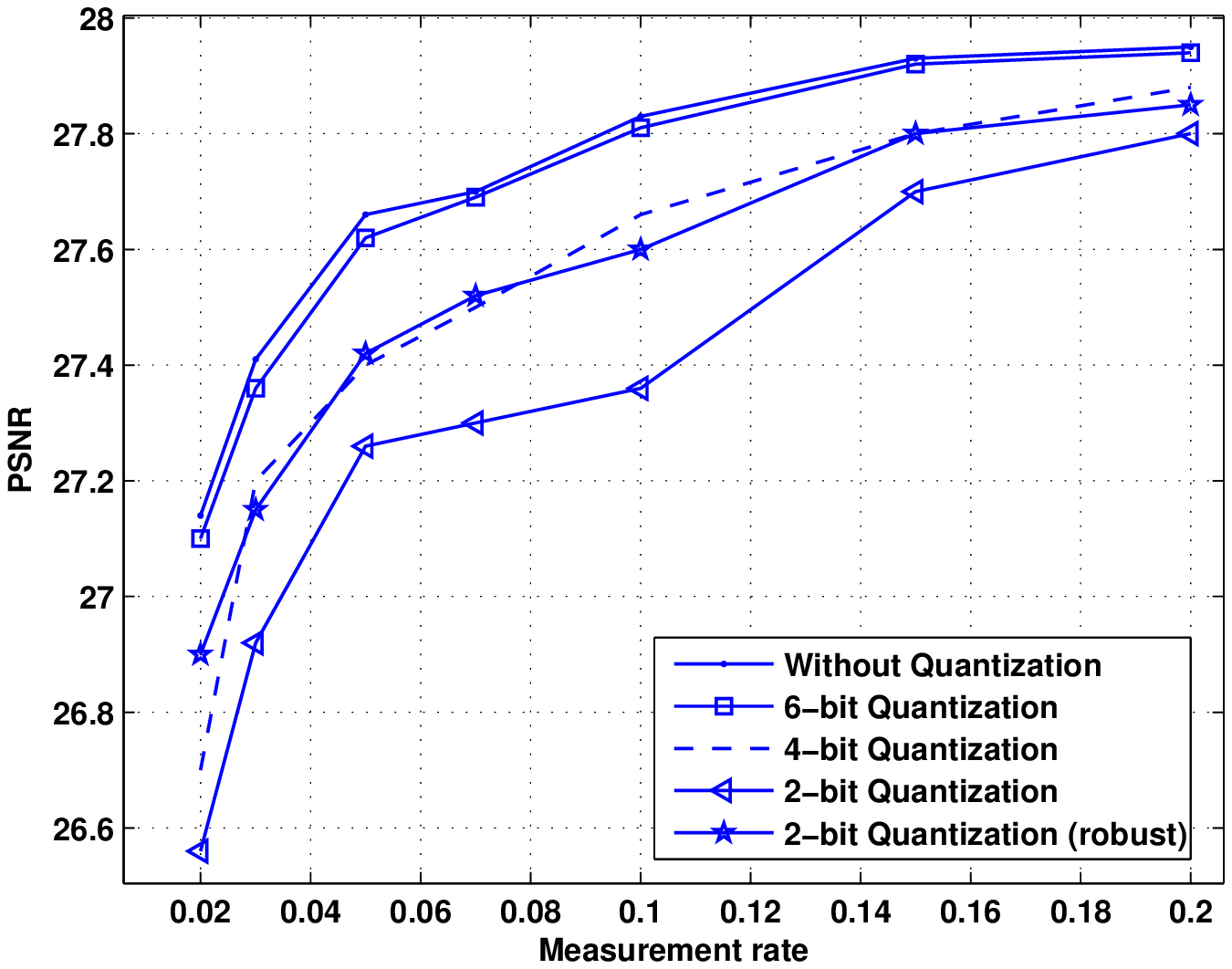} & \epsfxsize=2.8in \epsffile{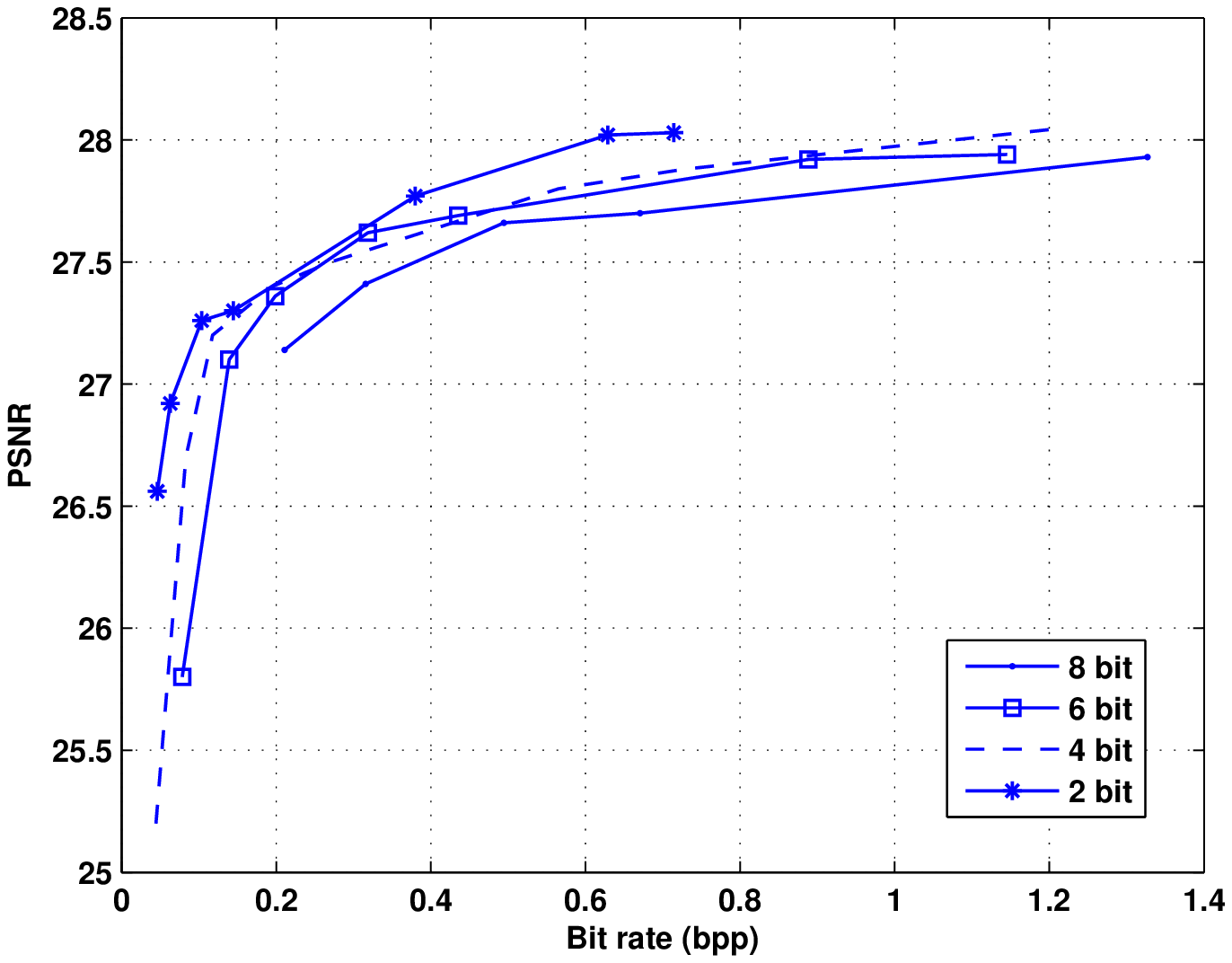} \\
   \mbox{(a) } & \mbox{(b)}
  \end{array}$
\caption{Effect of measurement quantization on the quality of the image $\hat{I}_2$ decoded with OPT-2 scheme in the Sawtooth dataset. The quality of the predicted image $\hat{I}_2$ is given in terms of (a) measurement rate and (b) bit rate. The benefit of using robust data cost is illustrated using a 2-bit uniform quantizer.  }
  \label{Fig:saw_effQuant_bits}
  \vspace{-0.1in}
  \end{figure*}

We now study the influence of the quantization bit rate on the RD performance of $\hat{I}_2$ with the OPT-2 optimization scheme. We compress the measurements $y_2$ using $2$-, $4$- and $6$-bits uniform quantizers.  As expected, the quality of the correlation estimation degrades when the number of bits reduces as shown in Fig.~\ref{Fig:saw_effQuant_bits}(a). However, it is largely compensated by the reduction in bit rate in the RD performance as confirmed by Fig.~\ref{Fig:saw_effQuant_bits}(b). This means that the proposed correlation estimation is relatively robust to quantization so that it is possible to attain good RD performance by drastic quantization of the measurements. Finally, we study the improvement offered by the robust data term $\tilde{E}_d$ (see Eq.~(\ref{eqn:robustdatacost})) in OPT-2, when the measurements have been compressed with a 2-bit uniform quantizer. From Fig. \ref{Fig:saw_effQuant_bits}(a) it is clear that the proposed robust data term improves the performance due to the efficient handling of noise in the quantized measurements.

\subsection{Multi-view image representation}
\begin{figure}
\centering
    \epsfxsize=2.7in \epsffile{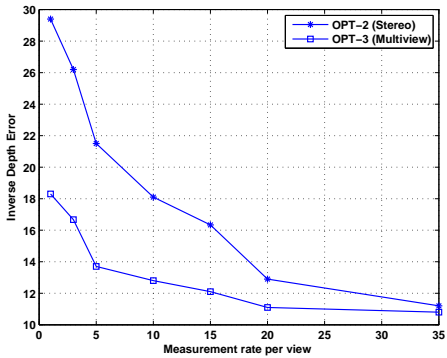}
 \caption{Inverse depth error at various measurement rates of the Tsukuba multi-view dataset. OPT-2  and OPT-3 problems are used to estimate the depth in stereo and multi-view scenarios respectively. The measurements are quantized using a 2-bit quantizer.}
   \label{Fig:tsu_multi-view_depth}
  \end{figure}

We finally evaluate the performance of our multi-view correlation estimation algorithms using five images from the \emph{Tsukuba} dataset (center, left, right, bottom and top views), and five frames (frames $3$-$7$) from the \emph{Flower Garden} sequence  \cite{multiview_gc}.  These datasets are down-sampled by a factor 2 and the resolution used in our experiments are of $144 \times 192$ and $120 \times 180$ pixels respectively. In both datasets, the reference image $I_1$ (center view and frame $5$ resp.) is encoded with a quality of approximately $33$~dB.  The measurements $½y_j, {\forall j \in \{ 1,2, 3, 4\}}$ computed from the remaining four images are quantized using a $2$-bit quantizer.  We first compare our results to a stereo setup where the disparity information is estimated with OPT-2 between the center and left images in Tsukuba dataset. Fig.~\ref{Fig:tsu_multi-view_depth} compares the inverse depth error (sum of the labels with error larger than one with respect to groundtruth) between the multi-view and stereo scenarios.  In this particular experiment, the parameters $\alpha_1$ and $\alpha_2$ are selected such that they minimize the error in the depth image with respect to the groundtruth. It is clear from the plot that the depth error is small for a given measurement rate when all the views are available. It should be noted that the $x$-axis in Fig.~\ref{Fig:tsu_multi-view_depth} represents the measurement rate per view. Hence, the total number of measurements used in the multi-view scenario is higher than for the stereo case.  However, these experiments show that the proposed multi-view scheme gives a better depth image when more images are available. Similar experimental findings have been observed for the Flower Garden sequence. 

\begin{figure}
\centering
    \epsfxsize=2.7in \epsffile{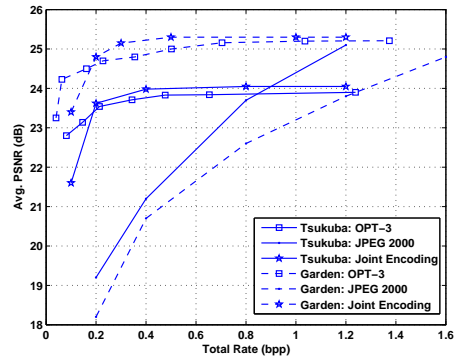}
 \caption{Comparison of the overall RD performances between the proposed OPT-3 scheme, joint encoding scheme and independent coding scheme based on JPEG~2000. Bit rate of the reference image $I_1$ is not included in the total bit budget.}
  \label{Fig:rdplot_multi-view}
  \end{figure}

We then study the RD performance of the proposed multi-view scheme in the decoding of four images (top, left, right, bottom images in the Tsukuba and frames $3, 4, 6, 7$ in the Flower Garden). The images are decoded by warping the reference image $\hat{I}_1$ using the estimated depth image. Fig.~\ref{Fig:rdplot_multi-view} compares the overall RD performance (for 4 images) of our multi-view scheme with respect to independent coding performance based on JPEG~2000. As expected, the proposed multi-view scheme outperforms independent coding solutions based on JPEG~2000 as it benefits from the correlation between images. Furthermore, as observed in distributed stereo coding the proposed multi-view coding scheme saturates at high rates, as the warping operator captures only the geometry and correlation between images and not the texture information.

Finally, we compare our results with a joint encoding approach where the depth image is estimated from the original images and transmitted to the joint decoder. At the decoder, the views are predicted from the reconstructed reference image $\hat{I}_1$ and the compressed depth image with the help of view prediction. The results are presented in Fig.~\ref{Fig:rdplot_multi-view} (denoted as \emph{Joint Encoding}), where the bit rate is computed only on the depth image encoded using a JPEG~2000 coding solution. The main difference between the proposed and joint encoding frameworks is that the quantized linear measurements are transmitted for a depth estimation in the former scheme, while the depth information is directly transmitted in the latter scheme. Therefore, by comparing these two approaches we can judge the accuracy of the estimated correlation model or equivalently the quality of the predicted view at a given bit rate.  From Fig.~\ref{Fig:rdplot_multi-view} we see that at low bit rate $< 0.2$, the proposed scheme estimates a better structural information compared to the joint encoding scheme, thanks to the geometry-based correlation representation. However at rates above $0.2$, we see that our scheme becomes comparable with joint coding solutions. This leads to the conclusion that the proposed scheme effectively estimates the depth information from the highly compressed quantized measurements. It should be noted that in joint encoding framework the depth images are estimated at a central encoder. In contrary to this, we estimate the depth images at the central decoder from the independently compressed visual information; this advantageously reduces the complexity at the encoder which makes it attractive for distributed processing applications.

\section{Conclusions} \label{sec:conc}
In this paper, we have presented a novel framework for the distributed representation of correlated images with quantized linear measurements, along with joint decoding algorithms that exploit the geometrical correlation among multiple images. We have proposed a regularized optimization problem in order to identify the geometrical transformations between compressed images, which result in smooth disparity or depth fields between a reference and one or more predicted image(s). We have proposed a low complexity algorithm for the correlation estimation problem which offers an effective trade-off between the complexity and accuracy of the solution. In addition, we have proposed a new consistency criteria such that transformations are consistent with the compressed measurements in the predicted image. Experimental results demonstrate that the proposed methodology provides a good estimation of dense disparity/depth fields in different multi-view image datasets. We also show that our geometry-based correlation model is more efficient than block-based correlation models. Finally, the consistent constraints prove to offer effective decoding quality such that the proposed algorithm outperforms JPEG~2000 and DSC schemes in terms of rate-distortion performance, even if the images are reconstructed by warping. This clearly positions our scheme as an effective solution for distributed image processing with low encoding complexity.

\bibliographystyle{IEEEtran}
\bibliography{cs_tip.bbl}
\end{document}